\documentclass[11pt]{article}
\usepackage[margin=1in]{geometry}
\usepackage{setspace}
\usepackage{graphicx}
\usepackage{amsmath}
\usepackage{amssymb}
\usepackage{booktabs}
\usepackage{hyperref}
\usepackage{xcolor}
\usepackage{listings}
\usepackage[numbers]{natbib}
\usepackage{comment}
\usepackage{siunitx}
\usepackage[margin=5pt,font=footnotesize]{subfig}

\hypersetup{
    colorlinks,
    linkcolor={red!50!black},
    citecolor={blue!50!black},
    urlcolor={blue!80!black}
}

\definecolor{light-gray}{gray}{0.95}
\newcommand{\doc}[1]{\colorbox{light-gray}{#1}}

\newcommand{\flip}[1]{\reflectbox{\rotatebox[origin=c]{180}{#1}}}

\begin{document}

\title{fastMRI: An Open Dataset and Benchmarks for Accelerated MRI}

\author{Jure Zbontar$^*$$^{1}$, Florian Knoll$^*$$^{2}$,  Anuroop Sriram$^*$$^{1}$, Tullie Murrell$^{1}$, Zhengnan Huang$^{2}$,
\\Matthew J. Muckley$^{2}$, Aaron Defazio$^{1}$, Ruben Stern$^{2}$, Patricia Johnson$^{2}$,\\
Mary Bruno$^{2}$, Marc Parente$^{2}$, Krzysztof J. Geras$^{2,3}$, Joe Katsnelson$^{2}$, Hersh Chandarana$^{2}$,\\
Zizhao Zhang$^{4}$, Michal Drozdzal$^{1}$, Adriana Romero$^{1}$, Michael Rabbat$^{1}$, Pascal Vincent$^{1}$,\\
Nafissa Yakubova$^{1}$, James Pinkerton$^{1}$, Duo Wang$^{2}$, Erich Owens$^{1}$,\\ C. Lawrence Zitnick$^{1}$, Michael P. Recht$^{2}$, Daniel K. Sodickson$^{2}$, Yvonne W. Lui$^{2}$}

\date{}
\maketitle
\vspace{-0.8cm}
\begin{center}
\begin{tabular}{l}
$^{1}$ Facebook AI Research\\
$^{2}$ NYU School of Medicine\\
$^{3}$ NYU Center for Data Science\\
$^{4}$ University of Florida\\
$^{*}$ denotes equal contribution
\end{tabular}

\medskip
Contact: \href{mailto:fastmri@med.nyu.edu}{fastmri@med.nyu.edu}, \href{mailto:fastmri@fb.com}{fastmri@fb.com}
\end{center}

\begin{abstract}
Accelerating Magnetic Resonance Imaging (MRI) by taking fewer measurements has the potential to reduce medical costs, minimize stress to patients and make MRI possible in applications where it is currently prohibitively slow or expensive. We introduce the fastMRI dataset, a large-scale collection of both raw MR measurements and clinical MR images, that can be used for training and evaluation of machine-learning approaches to MR image reconstruction. By introducing standardized evaluation criteria and a freely-accessible dataset, our goal is to help the community make rapid advances in the state of the art for MR image reconstruction. We also provide a self-contained introduction to MRI for machine learning researchers with no medical imaging background.
\end{abstract}

\section{Introduction}
\label{sec:intro-main} The excellent soft tissue contrast and flexibility of \emph{magnetic resonance imaging} (MRI) makes it a very powerful diagnostic tool for a wide range of disorders, including neurological, musculoskeletal, and oncological diseases. However, the long acquisition time in MRI, which can easily exceed 30 minutes, leads to low patient throughput, problems with patient comfort and compliance, artifacts from patient motion, and high exam costs. 

As a consequence, increasing imaging speed has been a major ongoing research goal since the advent of MRI in the 1970s. Increases in imaging speed have been achieved through both hardware developments (such as improved magnetic field gradients) and software advances (such as new pulse sequences).  One noteworthy development
in this context is parallel imaging, introduced in the 1990s, which 
allows multiple data points to be sampled simultaneously, rather than in a traditional sequential order
\citep{sodickson1997simultaneous,pruessmann1999sense,Griswold2002}. 

The introduction of \emph{compressed sensing} (CS) in 2006 \citep{candes-cs, Lustig2007} promised another breakthrough in the reduction of MR scan time. At their core, CS techniques speed up the MR acquisition by acquiring less measurement data than has previously been required to reconstruct diagnostic quality images.
Since undersampling of this kind violates the Nyquist-Shannon sampling theorem, aliasing artifacts are introduced which must be eliminated in the course of image reconstruction.
This can be achieved by incorporating additional \textit{a priori} knowledge during the image reconstruction process.

The last two years have seen the rapid development of machine learning approaches for MR image reconstruction, which hold great promise for further acceleration of MR image acquisition \citep{Hammernik2016,WangSYPZLFL16,varnet,SchlemperCHPR17,Zhu2018}. Some of the first work on this subject was presented at the 2016 annual meeting of the \emph{International Society for Magnetic Resonance in Medicine} (ISMRM). The 2017 ISMRM annual meeting included, for the first time, a dedicated session on machine learning for image reconstruction, and presentations on the subject at the 2018 annual meeting spanned multiple focused sessions, including a dedicated category for abstracts.

Despite this substantial increase in research activity, the field of MR image reconstruction still lacks large-scale, public datasets with consistent evaluation metrics and baselines. Many MR image reconstruction studies use datasets that are not openly available to the research community. This makes it challenging to reproduce and validate comparisons of different approaches, and it restricts access to work on this important problem to researchers associated with or cooperating with large academic medical centers where such data is available. 

In contrast, research in computer vision applications such as object classification has greatly benefited from the availability of large-scale datasets associated with challenges such as the \emph{ImageNet Large Scale Visual Recognition Challenge} (ILSVRC) \citep{Russakovsky2015}. Such challenges have served as a catalyst for the recent explosion in research activity on deep learning \citep{LeCun2015}. 

The goal of the fastMRI dataset is to provide a first step towards enabling similar breakthroughs in the machine-learning-based reconstruction of accelerated MR images. In this work we describe the first large-scale release of raw MRI data that includes 8344 volumes, consisting of 167,375 slices\footnote{A slice corresponds to one image.}, associated with \emph{in vivo} examinations from a range of MRI systems. In addition, we are releasing processed MR images in DICOM format from 20,000 knee and brain examinations from a representative clinical patient population, consisting of more than 1.57 million slices.

Prior to providing details about the dataset and about target reconstruction tasks with associated benchmarks, we begin with a brief primer on MR image acquisition and reconstruction, in order to enable non-MRI-experts to get up to speed quickly on the information content of the dataset.  In general, both the fastMRI dataset and this paper aim to connect the data science and the MRI research communities, with the overall goal of advancing the state of the art in accelerated MRI. 

\section{Introduction to MR Image Acquisition and Reconstruction}
\label{sec:intro} 
\begin{figure}
\centering
\includegraphics[height=5cm,keepaspectratio]{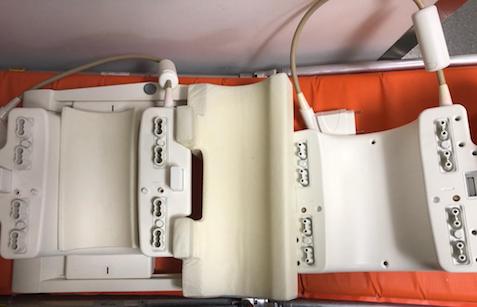}
\hspace{1cm}\includegraphics[height=5cm,keepaspectratio]{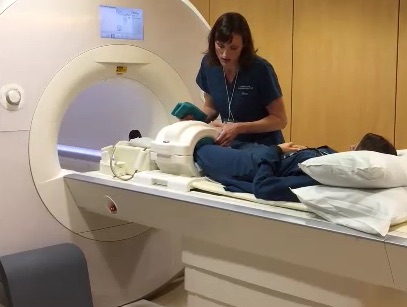} 
\caption{The receiver coil housing and its positioning on a patient for a knee MR examination.\label{fig:coils}}
\end{figure}

MR imaging is an indirect process, whereby cross-sectional images of the subject's anatomy are produced from frequency and phase measurements instead of direct, spatially-resolved measurements. A measuring instrument, known as a
\textit{receiver coil}, is placed in proximity to the area to be imaged (Figure~\ref{fig:coils}). During imaging, a sequence of spatially- and temporally-varying magnetic fields, called a ``pulse sequence,'' is applied by the MRI machine. This induces the body to emit resonant electromagnetic response fields which are measured by the receiver coil. 
The measurements typically correspond to points along a prescribed path through the multidimensional Fourier-space representation of an imaged body.
This Fourier space is known as \textit{k-space} in the medical imaging community. In the most basic usage of MR imaging, the full
Fourier-space representation of a region is captured by a sequence of samples that tile the space up to a specified maximum frequency. 

The spatially-resolved image $m$ can be estimated from the full k-space $y$ by performing an inverse multidimensional
Fourier transform:
\begin{equation}
	\hat{m} = \mathcal{F}^{-1}(y),
\end{equation}
where $\hat{m}$ is a noise-corrupted estimate of the true image $m$.

The number of samples captured in k-space is a limiting factor 
for the speed of MR imaging. Fewer samples can be captured by sampling up to a lower maximum frequency, however this produces images of lower spatial resolution. An alternative undersampling approach involves omitting some number of k-space samples within a given maximum frequency range, which then results in aliasing artifacts.  In order to remove these artifacts and infer the true underlying spatial structure of the imaged subject, one may apply a number of possible reconstruction strategies.

\begin{figure}
\centering
\newcommand{\multicoil}[1]{\includegraphics[width=0.332\textwidth,keepaspectratio]{#1}}
\newcommand{\multicoilhalf}[1]{\includegraphics[width=0.25\textwidth,keepaspectratio]{#1}}
\subfloat[k-space data from 15 coils]{\multicoil{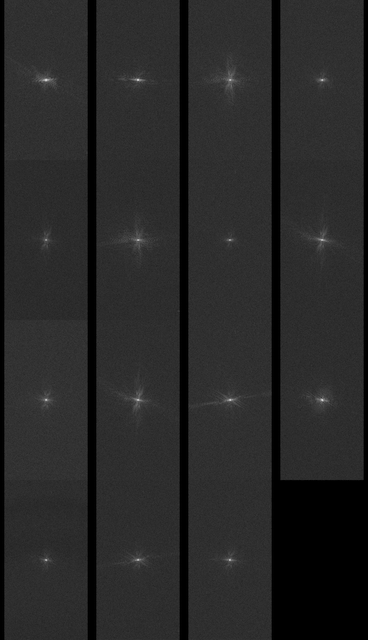}}
\subfloat[Individual coil spatial images from fully sampled data\label{fig:multi-coil-ifft-fully}]{\multicoil{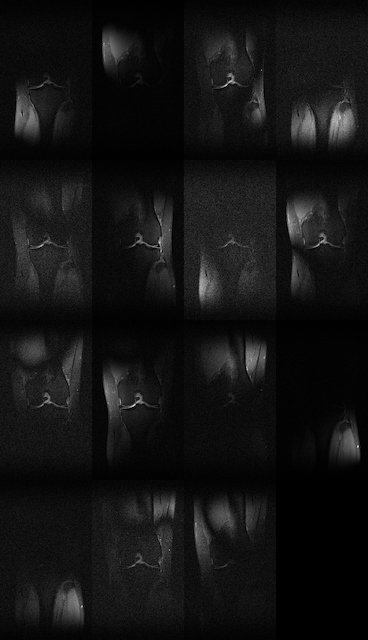}}
\subfloat[Coil sensitivity map magnitudes given by ESPIRiT]{\multicoil{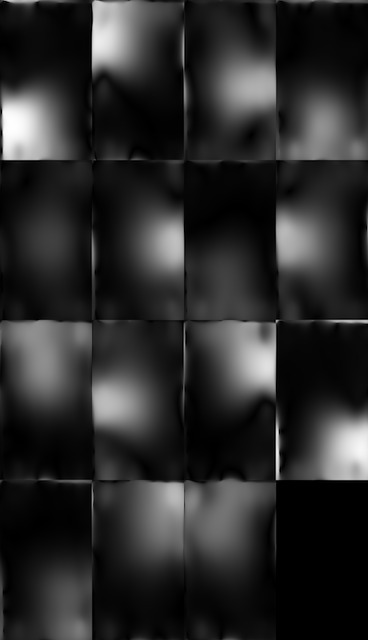}}

\subfloat[Ground truth, cropped to central region and vertically flipped]{\flip{\multicoilhalf{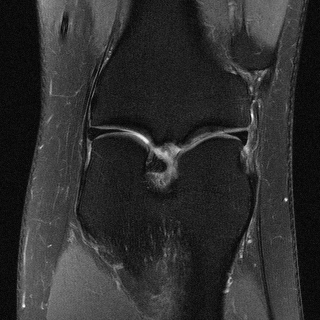}}}
\subfloat[Unregularized reconstruction]{\flip{\multicoilhalf{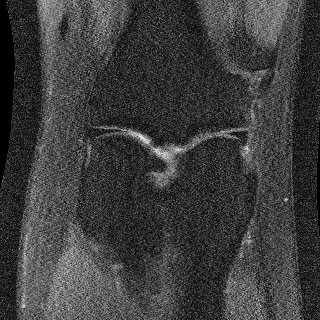}}}
\subfloat[Total variation penalty reconstruction]{\flip{\multicoilhalf{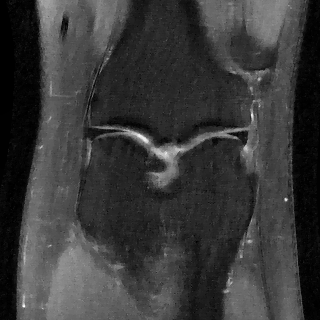}}}
\subfloat[Baseline model reconstruction]{\flip{\multicoilhalf{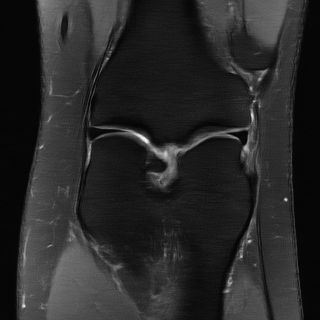}}}
\caption{Multi-coil MRI reconstruction \label{fig:multicoil-example}}
\end{figure}

\subsection{Parallel MR Imaging}
\label{sec:parallel-intro} In parallel MR imaging, multiple receiver coils are used, each of which produces a separate k-space measurement matrix. Each of these matrices is different, since the view each coil provides of the imaged volume is modulated by the differential  sensitivity that coil exhibits to MR signal arising from different regions.  In other words, each coil measures Fourier components of the imaged volume multiplied by a complex-valued position-dependent coil sensitivity map $S_i$.
The measured k-space signal $y_i$ for
coil $i$ in an array of $n_c$ coils is given by
\begin{equation}
	y_i = \mathcal{F}\left(S_{i}m\right)+\mathrm{noise},
\label{eq:parallel}
\end{equation}
where the multiplication is entry-wise. This is illustrated in Figure \ref{fig:multi-coil-ifft-fully}, which shows the absolute value of the inverse \emph{discrete Fourier transform} (DFT) of fully-sampled complex-valued k-space signals for each coil in a 15-element coil array. Each coil is typically highly sensitive in one region, and its sensitivity falls off significantly in other regions.

If the sensitivity maps are known, and the k-space sampling is full (i.e., satisfying the Nyquist sampling condition), then the set of linear relations between $m$ and each $y_i$ defines a linear system that is overdetermined by a factor of $n_c$. It may be inverted using a pseudoinverse operation to produce a reconstruction of $m$, as long as the linear system is full rank. The quality of this reconstruction will depend on the measurement noise, since the signal-to-noise ratio is poor in parts of the volume where the coil sensitivity is low.

In \emph{accelerated} parallel imaging, each coil's k-space signal is undersampled.
As long as the total number of measurements across all coils exceeds the number of image voxels to be reconstructed, an unregularized least squares solution
can still be used, leading to a theoretical $n_c$-fold speedup over fully-sampled single-coil imaging. Each extra coil effectively produces an additional ``sensitivity-encoded'' measurement of the volume \citep{pruessmann1999sense}, which augments the frequency and phase encoded measurements obtained from the sequential application of magnetic field gradients in the MR pulse sequence.  Estimates of coil sensitivity patterns, required for inversion of the undersampled multi-coil linear system, may be generated from separate low-resolution calibration scans. They may also be derived directly from the k-space measurements by fully sampling a comparatively small central region of k-space, which corresponds to low spatial frequencies.

In practice, the use of sub-sampling results in significant amplification of noise, and regularization is usually needed. In cases where a tight imaging field of view is used, or at imaging depths exceeding the dimensions of the individual coils, the sensitivity patterns of different coils spread out, thereby lowering the effective rank of the linear system, increasing noise amplification associated with the inverse operation, and limiting the maximum practical acceleration. As a result, in the clinic, parallel imaging acceleration factors are typically on the order of two to three. 


\subsection{Machine Learning Reconstruction of Undersampled MRI Data}
Classical approaches to MRI reconstruction solve a regularized inverse optimization problem to find the spatially-resolved image from the sub-sampled k-space data, in both 
the single-coil and the multi-coil case. We describe the classical approach in more detail in Section \ref{sec:model}.
In the machine learning approach, a reconstruction function
\begin{equation}
\hat{m} = B(y)
\end{equation}
is learned from input and output pair tuples $(y,m)$ drawn from a population. The goal is to find a function $B$ that minimizes the risk (i.e., expected loss) over the population distribution:
\begin{gather*}
B^{*}=\arg\min_{B}R(B),\\
\text{where } R(B)=\mathbb{E}_{(y,m)}\left[L\left(B\left(y\right),m\right)\right].
\end{gather*}
We discuss error metrics that may be used as loss functions $L$ in Section~\ref{sec:metrics}. In practice this optimization problem must be approximated with the empirical risk using a sample $\{(m^{(i)}, y^{(i)})\}_{i=1}^{n_{\text{data}}}$ from the population, with respect to a loss function $L$:
\begin{equation}
R_{\text{empirical}}(B)=\frac{1}{n_{\text{data}}}\sum_{i=1}^{n_{\text{data}}}L\left(B\left(y^{(i)}\right),m^{(i)}\right).
\end{equation}

\section{Prior Public Datasets}
\label{sec:relatedwork} \label{sec:related-datasets} The availability of public datasets has played an important role in advancing research in medical imaging, providing benchmarks to compare different approaches and leading to more impactful contributions. Early works such as DDSM \cite{DDSM}, SLIVER07 \cite{Heimann09} and CAUSE07 \cite{Ginneken07} triggered increasing efforts to collect new larger-scale biomedical datasets, which resulted in over one hundred public releases (counting the entries on \url{https://grand-challenge.org/}) to advance medical image analysis research. The vast majority of these datasets, which include a range of medical imaging modalities, are designed to test the limits of current methods in the tasks of segmentation, classification, and detection. Datasets such as BraTS \cite{BRATS}, LUNA \cite{LUNA}, ChestX-ray \cite{CHESTXRAY}, DeepLesion \cite{DEEPLESION}, and Camelyon \cite{CAMELYON}, UK biobank\footnote{https://imaging.ukbiobank.ac.uk}, ADNI (Alzheimer’s Disease Neuroimaging Initiative)\footnote{http://adni.loni.usc.edu/about/\#core-container} and TCIA (The Cancer Imaging Archive)\footnote{https://www.cancerimagingarchive.net/}. However, none of the most prominent public MRI datasets include k-space imaging data.

However, the current lack of large-scale reference standards for MR image reconstruction hinders progress in this important area. Most research uses synthetic k-space data that is
not directly acquired but rather obtained from post-processing of already-reconstructed
images \cite{UlHassanDarC17, ShitritR17, YangSLX17, YangSLX16, QuanJ16}. Research using small-scale proprietary raw k-space datasets is also common
\cite{HuangPCDHZ13, SchlemperCHPR17_dynamic, SchlemperCHPR17, Sandino2017DeepCN, Lonning2018}. 



In order to address the above-mentioned shortcomings, recent efforts have been devoted to collecting and publicly releasing datasets containing raw (unprocessed) k-space data; see, e.g., \cite{Sawyer_creationof, varnet}. However, the size of these existing datasets remains small. As an example, Table \ref{tab:datasets2} lists publicly available knee MR datasets containing raw k-space data. Although datasets such as these provide a valuable test bed for signal processing methods, larger datasets encompassing different anatomy are required to fully realize the potential of deep learning. 



\begin{table}
\begin{center}
\scriptsize
\begin{tabular}{lrll} \toprule
Dataset & Volumes & Body part & MR scan type
\\ \midrule
\href{http://mridata.org/}{NYU dataset} \cite{varnet} & 100 & knee & PD, T2  \\
\href{http://mridata.org/}{Stanford dataset 2D FSE} & 89 & knee &  PD  \\
\href{http://mridata.org/}{Stanford dataset 3D FSE} \cite{Sawyer_creationof} & 20 & knee & PD  \\
\href{http://mridata.org/}{Stanford undersampled dataset} & 38 & knee & PD  \\ \midrule
\textbf{fastMRI dataset} & \textbf{1594} & knee & PD \\
\bottomrule
\end{tabular}
\caption{\label{tab:datasets2} Publicly available MRI datasets containing k-space data}
\end{center}
\end{table}

\section{The fastMRI Dataset and Associated Tasks}
\label{sec:data} The fastMRI dataset (\href{http://fastmri.med.nyu.edu/}{http://fastmri.med.nyu.edu/})
contains four types of data from MRI acquisitions of knees and brains
\begin{description}
\item [Raw multi-coil k-space data:] unprocessed complex-valued multi-coil MR measurements.
\item [Emulated single-coil k-space data:] combined k-space data derived from multi-coil k-space data in such as way as to approximate 
single-coil acquisitions, for evaluation of single-coil reconstruction algorithms.
\item [Ground-truth images:] real-valued images reconstructed from 
\emph{fully-sampled} multi-coil acquisitions using the simple root-sum-of-squares method detailed below. These may be used as references to evaluate the quality of reconstructions.
\item [DICOM images:] spatially-resolved images for which the raw data was 
discarded during the acquisition process. These images are provided to represent a
larger variety of machines and settings than are present in the raw data.
\end{description}
This data was designed to enable two distinct types of \emph{tasks:}
\begin{enumerate}
\item {\bf Single-coil reconstruction task:} reconstruct images approximating the ground-truth from undersampled single-coil data.
\item {\bf Multi-coil reconstruction task:} reconstruct images approximating the ground-truth from undersampled multi-coil data.
\end{enumerate}

For each task we provide an official split of the k-space data and ground-truth images into \emph{training} and \emph{validation} subsets that contain fully-sampled acquisitions, 
as well as \emph{test} and \emph{challenge} subsets which contain k-space data that have been subjected to undersampling masks as described below. Ground-truth images are not being released for the test and challenge datasets. During training of a machine-learning model, the training k-space data should be programmatically masked following the same procedure.
The challenge subsets are not being released at the time of writing and are reserved for future challenges associated with the fastMRI dataset.

The rationale for having a single-coil reconstruction task (and for providing simulated single-coil data), even though reconstruction from multi-coil data is expected to be more precise, is twofold: (i) to lower the barrier of entry for researchers who may not be 
familiar with MRI data, since the use of a single coil removes a layer of complexity, and (ii) to include a task that is relevant for the single-coil MRI machines still in use throughout the world.

The DICOM images may be useful as additional data for training. Their distribution is different from that of the ground-truth images, since they were acquired with a larger diversity of scanners, manners of acquisition, reconstruction methods, and post-processing algorithms, so the application of transfer-learning techniques may be necessary.
Most DICOM images are the result of accelerated parallel imaging acquisitions and corresponding reconstructions, with image quality that differs from that of putative fully-sampled acquisitions and reconstructions. The ground-truth images may, in many cases, represent a higher standard of image quality than the clinical gold standard, for which full sampling is not routine or even practical.

\subsection{Anonymization}
\label{sec:dataset-inclusion} Curation of the datasets described here was part of a study approved by the NYU School of Medicine Institutional Review Board. Raw data was anonymized via conversion to the vendor-neutral ISMRMRD format~\cite{inati2017ismrm}. DICOM data was anonymized using the RSNA clinical trial processor. We performed manual inspection of each DICOM image for the presence of unexpected protected health information (PHI), manual checking of metadata in raw data files, as well as spot checking of all metadata and image content.

\subsection{Knee k-space Data}
\label{sec:raw-multi-coil-description} Multi-coil raw data was stored for 1,594 scans acquired for the purpose of diagnostic knee MRI. For each scan, a single fully sampled MRI volume was acquired on one of three clinical 3T systems (Siemens Magnetom Skyra, Prisma and Biograph mMR) or one clinical 1.5T system (Siemens Magnetom Aera). Data acquisition used a 15 channel knee coil array and conventional Cartesian 2D TSE protocol employed clinically at NYU School of Medicine. The dataset includes data from two pulse sequences, yielding coronal proton-density weighting with (PDFS, 798 scans) and without (PD, 796 scans) fat suppression (see Figure~\ref{fig:acquisition}). Sequence parameters are, as per standard clinical protocol, matched as closely as possible between the systems.
\begin{table}
\centering
\begin{tabular}{cc}\toprule
System & Number of scans \\ \midrule
Skyra 3T & 663 \\
Prisma 3T & 83 \\
Biograph mMR 3T & 153 \\
Aera 1.5T & 695 \\ \bottomrule
\end{tabular}
\caption{\label{tab:rawdata} Number of scans of knee raw data per scanner}
\end{table}
The following sequence parameters were used: Echo train length 4, matrix size $320 \times 320$, in-plane resolution $0.5$mm$\times 0.5$mm, slice thickness 3mm, no gap between slices. Timing varied between systems, with repetition time (TR) ranging between 2200 and 3000 milliseconds, and echo time (TE) between 27 and 34 milliseconds.

\subsection{Brain k-space Data}

Data from 6970  fully sampled brain MRIs were obtained using 11 magnets across 5 clinical locations using 1.5T and 3T field strengths. Magnets include the 3T Prisma, Skyra, Biograph, Tim Trio and the 1.5T Avanto and Aera Magnetom (Siemens Healthcare, Erlangen Germany). The raw dataset includes axial T1 weighted, T2 weighted and FLAIR images. Some of the T1 weighted acquisitions included admissions of contrast agent (labelled T1 POST) (see Figure~\ref{fig:brain_acquisition}). Not all imaging volumes included all pulse sequences. The exact distribution of contrasts and field strengths is given in table \ref{tab:brain_rawdata}.

To ensure data de-identification, we used only axial 2-D images in this dataset. We used zero matrices to replace the k-space slices $\gtrapprox 5mm$ below the orbital rim. All processed k-spaces were then reconstructed to images in DICOM format, loaded into a picture archival communication system (PACS) and all images were visually checked by certified MR technologists to confirm exclusion of identifying facial features.

\begin{table}
\centering
\begin{tabular}{ccc}\toprule
Field Strength & 1.5T & 3T \\ \midrule
T1 & 375 & 407 \\
T1 POST & 849 & 641 \\
T2 & 1651 & 2515 \\
FLAIR & 126 & 406\\ \midrule
Total & 3001 & 3969 
 \\ \bottomrule
\end{tabular}
\caption{\label{tab:brain_rawdata} Number of scans for the different contrasts and scanner field strengths of the brain raw dataset.}
\end{table}

\subsection{Knee emulated Single-coil k-Space Data}
\label{sec:esc}
We used an emulated single-coil (ESC) methodology to simulate single-coil data from a multi-coil acquisition \citep{tygert2018simulating}. ESC computes a complex-valued linear combination of the responses from multiple coils, with the linear combination fitted to the ground-truth root-sum-of-squares reconstruction in the least-squares sense.  

\subsection{Knee DICOM Data}
\label{sec:dicom_data}
\label{sec-dicom} In addition to the scanner raw data described above, the fastMRI dataset includes DICOM data from 10,000 clinical knee MRI scans. These images represent a wider variety of scanners and pulse sequences than those represented in the collection of raw data. Each MR exam for which DICOM images are included typically consisted of five clinical pulse sequences: 
\begin{enumerate}
\item Coronal proton-density weighting without fat suppression,
\item Coronal proton-density weighting with fat suppression,
\item Sagittal proton-density weighting without fat suppression,
\item Sagittal T2 weighting with fat suppression, and
\item Axial T2 weighting with fat suppression.
\end{enumerate}
The two coronal sequences have the same basic specifications (matrix size, etc) as the sequences associated with raw data. The sagittal and axial sequences have different matrix sizes and have no direct correspondence to the sequences represented in raw data.

The Fourier transformation of an image from a DICOM file does not directly correspond to the originally measured raw data, due to the inclusion of additional post-processing steps in the vendor-specific reconstruction pipeline. Most of the DICOM images are also derived from accelerated acquisitions and are reconstructed with parallel imaging algorithms, since this baseline acceleration represents the current clinical standard.  The image quality of DICOM images, therefore, is not equivalent to that of the ground truth images directly associated with fully sampled raw data. The DICOM images are distinct from the validation, test, or challenge sets.

\subsection{Brain DICOM}

10,000 brain MRI DICOM studies are also included. Axial 2D image volumes are included with the following pulse sequences: T1, T2, and T2 FLAIR. All studies represent unique individuals and there is no subject overlap with the brain rawdata. In addition to the deidentification procedures detailed above, the brain image volumes were cropped to exclude identifiable facial features, following which each image was visually inspected to confirm appropriate deidentification. Finally, we present 10,000 brain MRI DICOM studies from 10,000 unique subjects, each one including axial 2D DICOM image volumes through the majority of the brain representing a broad range of neurological pathologies. Not all studies include all pulse sequences.

\subsection{Ground Truth}
\label{sec:ground-truth} The root-sum-of-squares reconstruction method applied to the fully sampled k-space data \citep{root-sum-square} provides the ground truth for the multi-coil dataset. The single-coil dataset includes two ground truth reconstructions, which we denote ESC and RSS. The ESC ground truth is given by the inverse Fourier transform of the single-coil data, and the RSS ground truth is given by the root-sum-of-squares reconstruction computed on the multi-coil data that were used to generate the virtual single-coil k-space data. All ground truth images are cropped to the central $320 \times 320$ pixel region to compensate for readout-direction oversampling that is standard in clinical MR examinations. 

The root-sum-of-squares approach \citep{root-sum-square} is one of the most commonly-used coil combination methods in clinical imaging. It first applies the inverse Fourier Transform to the k-space data from each coil:
\begin{equation}
	\tilde{m}_i = \mathcal{F}^{-1}(y_i),
\end{equation}
where $y_i$ is the k-space data from the $i$th coil and $\tilde{m}_i$ is the $i$th coil image. Then, the individual coil images are combined voxel by voxel as follows:
\begin{equation}
\tilde{m}_{\text{rss}} = \left(\sum_{i=0}^{n_c} \left| \tilde{m}_i \right|^2\right)^{1/2},
\label{eq:RSS}
\end{equation}
where $\tilde{m}_{\text{rss}}$ is the final image estimate and $n_c$ is the number of coils. The root-sum-of-squares image estimate is known to converge to the optimal, unbiased estimate of $m$ in the high-SNR limit \citep{larsson2003snr}.
\begin{figure}
\centering
\subfloat[][]{\includegraphics[width=0.49\textwidth]{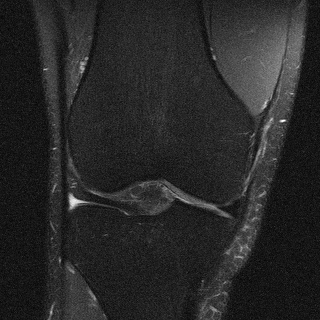}}
\subfloat[][]{\includegraphics[width=0.49\textwidth]{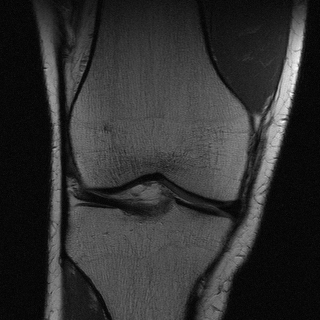}}
\caption{A proton-density weighted image (a) with fat suppression (PDFS) and (b) without fat suppression (PD). Fat has a high signal response in MR imaging, which can make details in other regions difficult to see. Fat-suppressed scans typically have higher noise.}
\label{fig:acquisition}
\end{figure}

\begin{figure}
\centering
\subfloat[][]{\reflectbox{\rotatebox[origin=c]{180}{
\includegraphics[width=0.49\textwidth]{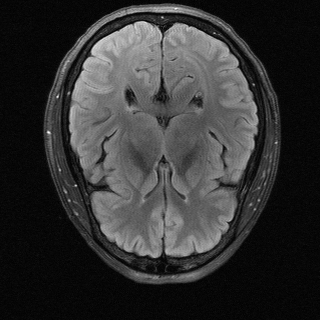}}}}
\subfloat[][]{\reflectbox{\rotatebox[origin=c]{180}{
\includegraphics[width=0.49\textwidth]{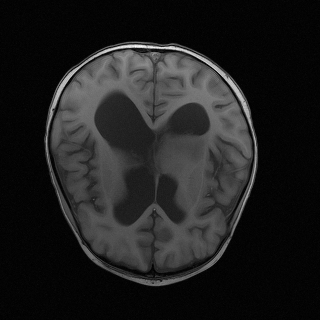}}}}
\\
\subfloat[][]{\reflectbox{\rotatebox[origin=c]{180}{\includegraphics[width=0.49\textwidth]{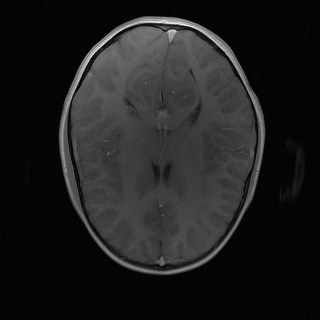}}}}
\subfloat[][]{\reflectbox{\rotatebox[origin=c]{180}{\includegraphics[width=0.49\textwidth]{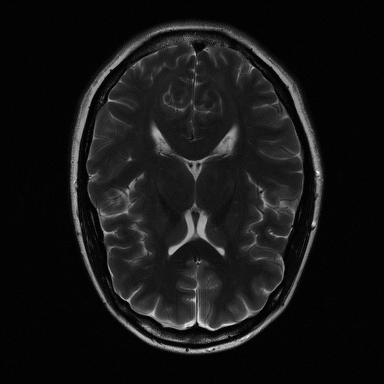}}}}
\caption{Axial brain MRI images with different contrasts: (a) FLAIR, (b) T1 weighted (c) T1 weighted with contrast agent (T1 POST), and (d) T2 weighted.}
\label{fig:brain_acquisition}
\end{figure}

\subsection{Dataset Split}
\begin{table}
\centering
\begin{tabular}{ccccc}\toprule
 & \multicolumn{2}{c}{Volumes} & \multicolumn{2}{c}{Slices} \\ 
 \cmidrule(r){2-3} \cmidrule(r){4-5}
& Multi-coil & Single-coil & Multi-coil & Single-coil \\ \midrule
training & 973 & 973 & 34,742 & 34,742 \\
validation & 199 & 199 & 7,135 & 7,135 \\
test & 118 & 108 & 4,092 & 3,903 \\
challenge & 104 & 92 & 3,810 & 3,305 \\ \bottomrule
\end{tabular} 
\caption{\label{tbl:datasets} Volumes and slices in each set}
\end{table}
\label{sec:splits} Each volume is randomly assigned to one of the following six component datasets: training, validation, multi-coil test, single-coil test, multi-coil challenge, or single-coil challenge. Table~\ref{tbl:datasets} shows the number of volumes assigned to each dataset. The training and validation datasets may be used to fit model parameters or to determine hyperparameter values. The test dataset is used to compare the results across different approaches. To ensure that models do not overfit to the test set, the ground truth reconstructions are not publicly released for this set. Evaluation on the test set is accomplished by uploading results to the public leaderboard at \href{http://fastmri.org/}{http://fastmri.org/}.  The challenge portion of the dataset will be forthcoming.

A volume from the train or validation dataset is used in both the single-coil and multi-coil tracks, whereas a volume from the test or challenge dataset is only used in either the single-coil or the multi-coil track. Volumes were only included in a single test or challenge set to ensure information from one could not be used to help the result in another.

\subsection{Cartesian Undersampling}
\label{sec:undersampling} 
Volumes in the test and challenge datasets contain undersampled k-space data. The undersampling is performed by retrospectively masking k-space lines from a fully-sampled acquisition. k-space lines are omitted only in the phase encoding direction, so as to simulate physically realizable accelerations in 2D data acquisitions. The same undersampling mask is applied to all slices in a volume, with each case consisting of a single volume. The overall acceleration factor is set randomly to either four or eight (representing a four-fold or an eight-fold acceleration, respectively), with equal probability for each.

\begin{figure}[t]
\hspace*{\fill} \subfloat[][\label{fig:4fold}Random mask with 4-fold acceleration]{\includegraphics[width=0.35\textwidth]{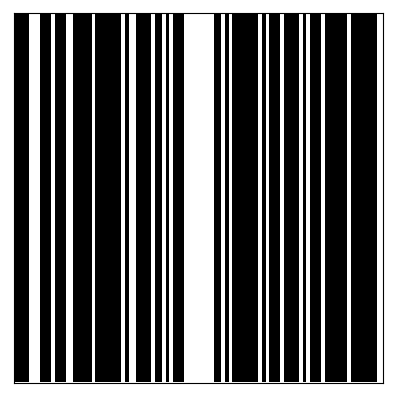}}\hfill \subfloat[][\label{fig:8fold}Random mask with 8-fold acceleration]{\includegraphics[width=0.35\textwidth]{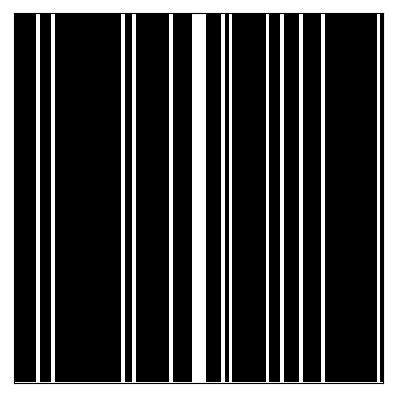}} \hspace*{\fill}
\\
\hspace*{\fill} \subfloat[][\label{fig:4fold_equi}Equispaced mask with 4-fold acceleration]{\includegraphics[height=0.32\textwidth,width=0.32\textwidth]{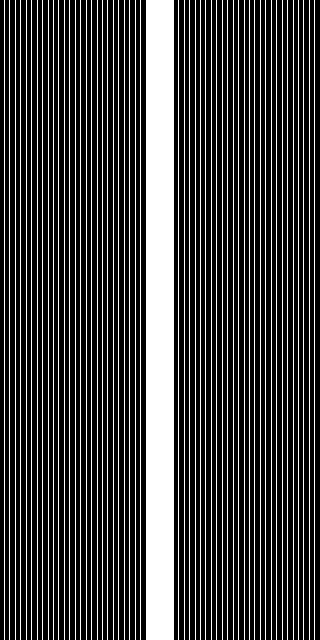}}\hfill 
\subfloat[][\label{fig:8fold_equi}Equispaced mask with 8-fold acceleration]{\includegraphics[height=0.32\textwidth,width=0.32\textwidth]{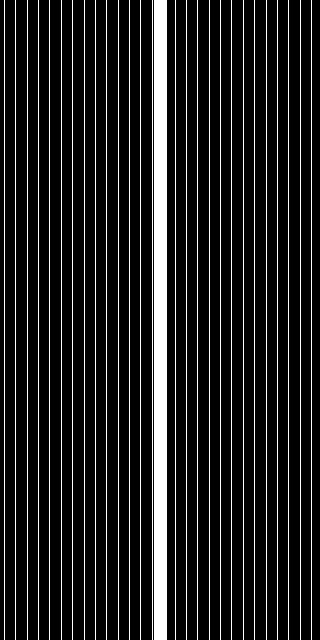}}\hspace*{\fill} 
\caption{\label{fig:kspace_mask} Examples of undersampled k-space trajectories}
\end{figure}

All undersampling masks are generated by first including some number of adjacent lowest-frequency k-space lines to provide a fully-sampled k-space region. When the acceleration factor equals four, the fully-sampled central region includes 8\% of all k-space lines; when it equals eight, 4\% of all k-space lines are included. The remaining k-space lines are included differently for both knee and brain cases. For knee, the remaining k-space lines are included uniformly at random, with the probability set so that, on average, the undersampling mask achieves the desired acceleration factor. Random undersampling is chosen in order to meet the general conditions for compressed sensing \citep{candes-cs,Lustig2007}, for a fair comparison of learned reconstruction algorithms with traditional sparsity-based regularizers. For brain, after a random offset from the start, the remaining lines are sampled equidistant from each other with a spacing that achieves the desired acceleration factor. Equidistant was chosen because of ease of implementation on existing MRI machines. Figure ~\ref{fig:kspace_mask} depicts the k-space trajectories for random and equidistant undersampling at four and eight acceleration factors.

\section{Metrics}
\label{sec:metrics} The assessment of MRI reconstruction quality is of paramount relevance to develop and compare machine learning and medical imaging systems \cite{wang2009mean,wang2004image,chandler2013seven,zhang2011fsim}. The most commonly used evaluation metrics in the MRI reconstruction literature \cite{chandler2013seven} include (normalized) mean squared error, which measures pixel-wise intensity differences between reconstructed and reference images, and signal-to-noise ratio, which measures the degree to which image information rises above background noise. These metrics are appealing because they are easy to understand and efficient to compute. However, they both evaluate pixels independently, ignoring the overall image structure.

Additional metrics have been introduced in the literature to capture structural distortion \cite{teo1994perceptual,eckert1998perceptual,zhang2011fsim}. For example, the structural similarity index ~\cite{wang2004image} and its extended version, multiscale structural similarity~\cite{wang2003multiscale}, provide a mechanism to assess the perceived quality of an image using local image patches. 

The most recent developments in the computer vision literature leverage pretrained deep neural networks to measure the perceptual quality of an image by computing differences at the representation level \cite{johnson2016perceptual}, or by means of a downstream task such as classification \cite{salimans2016improved}. 

In the remainder of this section, we review the definitions of the commonly-used metrics of normalized mean square error, peak signal-to-noise ratio, and structural similarity.  As is discussed later, while we expect these metrics to serve as a familiar starting point, we also hope that the fastMRI dataset will enable robust investigations into improved evaluation metrics as well as improved reconstruction algorithms.

\subsection{Normalized Mean Square Error}
\label{sec:mse} The \emph{normalized mean square error} (NMSE) between a reconstructed image or image volume represented as a vector $\hat{v}$ and a reference image or volume $v$ is defined as
\begin{equation}
\text{NMSE}(\hat{v}, v) = \frac{\| \hat{v} - v \|_2^2}{\| v \|_2^2},
\end{equation}
where $\| \cdot \|_2^2$ is the squared Euclidean norm, and the subtraction is performed entry-wise.  In this work we report NMSE values computed and normalized over full image volumes rather than individual slices, since image-wise normalization can result in strong variations across a volume.  

NMSE is widely used, and we recommend that it be reported as the primary measure of reconstruction quality for experiments on the fastMRI dataset. However, due to the many downsides of NMSE, such as a tendency to favor smoothness rather than sharpness, we recommend also reporting additional metrics such as those described below.

\subsection{Peak Signal-to-Noise Ratio}
\label{sec:psnr} The \emph{peak signal-to-noise ratio} (PSNR) represents the ratio between the power of the maximum possible image intensity across a volume and the power of distorting noise and other errors:
\begin{equation}
\text{PSNR}(\hat{v}, v) = 10\log_{10} \frac{\max(v)^2}{\text{MSE}(\hat{v}, v)}.
\end{equation}
Here $\hat{v}$ is the reconstructed volume, $v$ is the target volume, $\max(v)$ is the largest entry in the target volume $v$, $\text{MSE}(\hat{v}, v)$ is the mean square error between $\hat{v}$ and $v$ defined as $\frac{1}{n}\| \hat{v} - v \|_2^2$ and $n$ is the number of entries in the target volume $v$. Higher values of PSNR (as opposed to lower values of NMSE) indicate a better reconstruction.

\subsection{Structural Similarity}
\label{sec:ssim} The \emph{structural similarity} (SSIM) index measures the similarity between two images by exploiting the inter-dependencies among nearby pixels. SSIM is inherently able to evaluate structural properties of the objects in an image and is computed at different image locations by using a sliding window. The resulting similarity between two image patches $\hat{m}$ and $m$ is defined as
\begin{equation}
\text{SSIM}(\hat{m}, m) = \frac{(2 \mu_{\hat{m}} \mu_m + c_1)(2 \sigma_{\hat{m}m} + c_2)}{(\mu_{\hat{m}}^2 + \mu_m^2 + c_1)(\sigma_{\hat{m}}^2 + \sigma_m^2 + c_2)},
\end{equation}
where $\mu_{\hat{m}}$ and $\mu_m$ are the average pixel intensities in $\hat{m}$ and $m$, $\sigma_{\hat{m}}^2$ and $\sigma_m^2$ are their variances, $\sigma_{\hat{m}m}$ is the covariance between ${\hat{m}}$ and $m$ and $c_1$ and $c_2$ are two variables to stabilize the division; $c_1 = (k_1 L)^2$ and $c_2 = (k_2 L)^2$. For SSIM values reported in this paper, we choose a window size of $7 \times 7$, we set $k_1=0.01$, $k_2=0.03$, and define $L$ as the maximum value of the target volume, $L = \max(v)$.

\subsection{L1 Error}
\label{sec:l1} It is sometimes advantageous to use the L1 loss
\begin{equation}
\text{L}_1 (\hat{v}, v) = \| \hat{v} - v \|_1,
\end{equation}
for training machine learning models on computer vision tasks, even when evaluation is performed under L2 norm losses such as MSE \citep{l1}. The baseline models in Section \ref{sec:singlecoil_deep_baseline} were trained using L1 loss.

\section{Baseline Models}
\label{sec:model} 
Along with releasing the fastMRI data, we detail two reference approaches to be used as reconstruction baselines: a classical non-machine learning approach, and a deep-learning approach. Each of these baselines has versions tailored for single-coil or multi-coil data. The ``classical'' baselines are comprised of reconstruction methods developed by the MRI community over the last 30+ years. These methods have been extensively tested and validated, and many have demonstrated robustness sufficient for inclusion in the clinical workflow. By comparison, machine learning reconstruction methods are relatively new in MRI, and deep-learning reconstruction techniques in particular have emerged only in the past few years. We include some deliberately rudimentary deep-learning models as starting points, with the expectation that future learning algorithms will provide markedly improved performance.

\subsection{Single-coil Classical Baselines (knee only)}
\begin{figure}[t!]
\centering
\newcommand{\singlecoil}[1]{\includegraphics[width=0.25\textwidth,keepaspectratio]{#1}}
\subfloat[Cropped and vertically flipped reconstruction from fully sampled k-space data]{\flip{\singlecoil{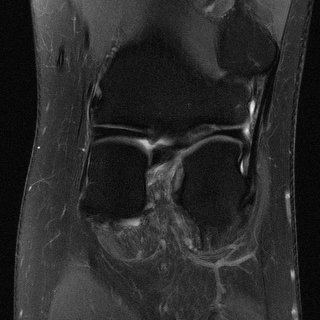}}}
\subfloat[Rectangular masked k-space]{\makebox[0.25\textwidth]{\includegraphics[height=0.25\textwidth,keepaspectratio]{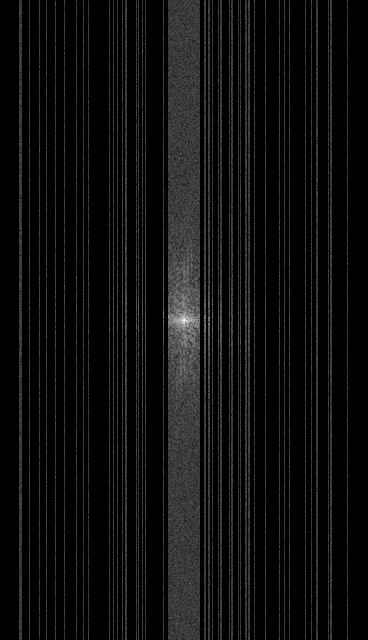}}} 
\subfloat[Reconstruction via zero-filled IFFT]{\flip{\singlecoil{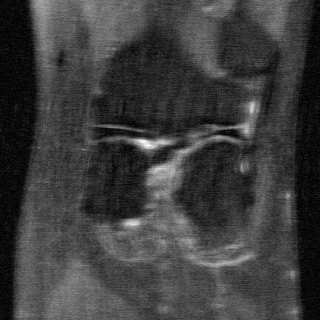}}}
\subfloat[Deep-learning baseline UNET reconstruction]{\flip{\singlecoil{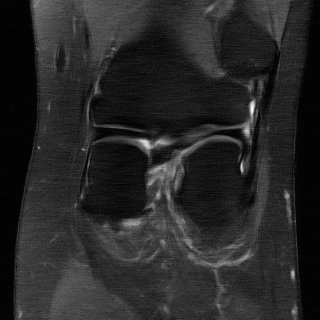}}}

\subfloat[Multiscale Daubechies discrete wavelet transform \label{fig:wavelet-transform}]{\singlecoil{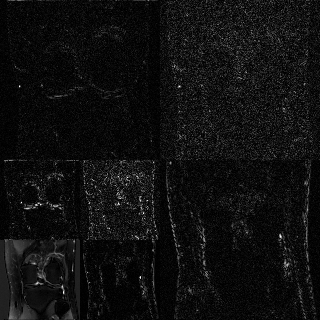}}
\subfloat[L1 Wavelet penalty reconstruction \label{fig:singlecoil-wavelet}]{\flip{\singlecoil{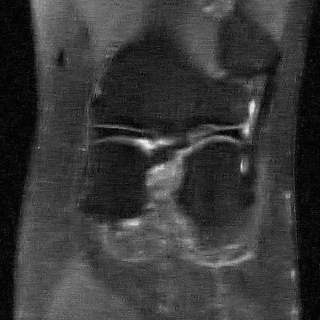}}}
\subfloat[Image gradients as given by a Sobel filter\label{fig:singlecoil-sobel}]{\flip{\singlecoil{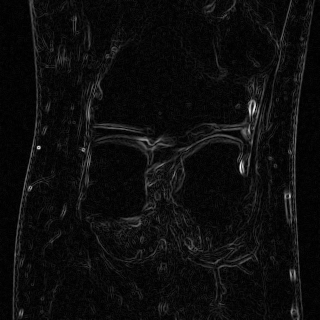}}}
\subfloat[Regularized total-variation reconstruction \label{fig:singlecoil-tv}]{\flip{\singlecoil{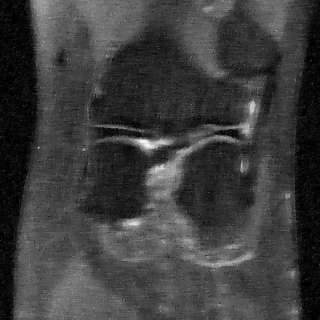}}}
\caption{Single-coil reconstruction}
\end{figure}
\label{sec:singlecoil_baseline} In the single-coil imaging setting, the task is to reconstruct an image, $m$, from k-space observations, $y$. In the presence of undersampling, the vector $y$ has a length smaller than that of $m$. Therefore there are, in principle, infinitely many possibilities for $m$ that can be mapped onto a single $y$. The advent of compressed sensing \citep{candes-cs,Lustig2007} provided a framework for reconstruction of images from undersampled data that closely approximate images derived from fully-sampled data, subject to sparsity constraints. Compressed sensing theory requires the images in question to be sparse in some transform domain. Two common examples are to assume sparsity in the wavelet domain, or to assume sparsity of the spatial gradients of the image. The particular assumption impacts the mathematical formulation of the reconstruction problem, either in the cost function or through a regularization term.

More concretely, the sparse reconstruction approach consists of finding an image $m$ whose Fourier space representation is close to the measured k-space matrix $y$ at all measured spatial frequencies, 
yet at the same time minimizes a sparsity-inducing objective $R(m)$ that penalizes unnatural reconstructions:
\newcommand{\minimize}{\mathop{\operatorname{minimize}}}
\begin{equation}
\minimize_{m}\;R\left(m\right)\;
\text{s.t.}\;\left\Vert \mathcal{P}\left(\mathcal{F}\left(m\right)\right) - y\right\Vert _{2}^{2}\leq\epsilon. \label{eq:cs-constraint}
\end{equation}
Here, \begin{math}\mathcal{P}\end{math} is a projection function that zeros out entries that are masked, and \begin{math}\epsilon\end{math} is a specified small threshold value. In most applications it is easier to work with a soft penalty instead of a constraint, so the Lagrangian dual form of Equation \ref{eq:cs-constraint} is used instead, with penalty parameter $\lambda$:
\begin{equation}
\minimize_{m}\;\frac{1}{2}\left\Vert \mathcal{P}\left(\mathcal{F}\left(m\right)\right)-y\right\Vert _{2}^{2}+\lambda R\left(m\right).
\label{eq:CS}
\end{equation}
For a convex regularizer $R$, there exists, for any choice $\epsilon>0$, a value $\lambda$ such that these two formulations have equivalent solutions.

The most common regularizers used for MRI are:
\begin{align*}
R_{L1}\left(m\right) & =\left\Vert m\right\Vert _{1},\\
R_{\text{wavelet}}\left(m\right) & =\left\Vert \Psi\left(m\right)\right\Vert _{1}\;\left(\Psi\text{ is a discrete wavelet transform}\right),\\
R_{TV}(m) & =\sum_{i,j}\sqrt{\left|m_{i+1,j}-m_{i,j}\right|^{2}+\left|m_{i,j+1}-m_{i,j}\right|^{2}}.
\end{align*}
The $L1$ penalty works best when the MR images are sparse in image space, for instance in vascular imaging (e.g., \citet{Yamamoto2018MagneticRA}). This is not the case for most MRI applications.
The total-variation (TV) penalty encourages sparsity in the spatial gradients of the reconstructed image, as given by a local finite-difference approximation \citep{rudin_tv} (Figure~\ref{fig:singlecoil-sobel}). The TV regularizer can be very effective for some imaging protocols, but it also has a tendency to remove detail (Figure~\ref{fig:singlecoil-tv}).
The $R_{\text{wavelet}}$ penalty encourages sparsity in the discrete wavelet transform of the image. Most natural images exhibit significant sparsity when expressed in a wavelet basis. The most commonly used transform is the Multiscale Daubechies (DB2) transform (Figure~\ref{fig:wavelet-transform}). To date, due to their computational complexity as well as their tendency to introduce compression artifacts or oversmoothing, compressed sensing approaches have taken some time to gain acceptance in the clinic, though commercial implementations of compressed sensing are currently beginning to appear.

The single-coil classical baseline provided with the fastMRI dataset was adopted from the widely-used open-source BART toolkit (Appendix \ref{appendix:bart-detail}), using total variation as the regularizer.  We ran the optimization algorithm for 200 iterations on each slice independently.

\begin{table}[htb]
\centering
\begin{tabular}{cccccccc}\toprule
\multicolumn{8}{c}{Single-coil classical baseline (TV model) applied to knee validation data} \\
\midrule
Acceleration & Regularization Weight & \multicolumn{2}{c}{NMSE} & \multicolumn{2}{c}{PSNR} & \multicolumn{2}{c}{SSIM} \\ 
 \cmidrule(r){3-4} \cmidrule(r){5-6} \cmidrule(r){7-8} 
& & PD & PDFS & PD & PDFS & PD & PDFS \\ \midrule
4-fold & $10^{-4}$ & 0.0355 & 0.0919 & 30.2 & 27.6 & 0.637 & \textbf{0.506} \\
 & $10^{-3}$ & 0.0342 & 0.0916 & 30.4 & 27.6 & 0.641 & 0.505 \\
 & $10^{-2}$ & \textbf{0.0287} & \textbf{0.09} & \textbf{31.4} & \textbf{27.7} & \textbf{0.645} & 0.494 \\
 & $10^{-1}$ & 0.0313 & 0.0993 & 30.9 & 27.3 & 0.575 & 0.399 \\
 & 1 & 0.0522 & 0.124 & 28.5 & 26.2 & 0.526 & 0.327 \\
\midrule
8-fold & $10^{-4}$ & 0.0708 & 0.118 & 27.1 & 26.4 & 0.551 & \textbf{0.417} \\
 & $10^{-3}$ & 0.0699 & 0.118 & 27.1 & 26.4 & 0.553 & 0.416 \\
 & $10^{-2}$ & 0.063 & \textbf{0.117} & 27.7 & 26.4 & \textbf{0.564} & 0.408 \\
 & $10^{-1}$ & \textbf{0.0537} & \textbf{0.117} & \textbf{28.4} & \textbf{26.5} & 0.55 & 0.357 \\
 & 1 & 0.0742 & 0.132 & 26.9 & 25.9 & 0.538 & 0.333 \\
\bottomrule
\end{tabular} 
\caption{\label{tbl:tv_single} Validation set results for the classical baseline model with Total Variation regularization for the single-coil task. Bold-faced numbers indicate the best performance for each image quality metric.}
\end{table}

Table \ref{tbl:tv_single} summarizes the results of applying this method to the single-coil validation data with different regularization strengths and different acceleration factors. These results indicate that NMSE and PSNR metrics are highly (inversely) correlated and generally favor models with stronger regularization than SSIM does. Stronger regularization generally results in smoother images that lack the fine texture of the ground truth images. A regularization parameter of 0.01 yields the best results for 4-fold acceleration in most cases, whereas the higher 8-fold acceleration gets slightly better results with a regularization parameter of 0.1.

\subsection{Multi-coil Classical Baselines}
\label{sec:multicoil_classical_baseline}
When multiple receiver coils are used, the reconstruction process must combine information from multiple channels into one image. Multi-coil acquisitions currently represent the norm in clinical practice, for two principal reasons: they provide increased SNR, as compared with single-coil acquisitions, over extended fields of view, and they enable acceleration via parallel imaging. Equation \ref{eq:parallel} in Section \ref{sec:parallel-intro} describes the forward model for parallel imaging.  The SENSE formulation \citep{pruessmann1999sense} of parallel image reconstruction involves direct inversion of this forward model, via a suitable pseudoinverse.  Leveraging the convolution property of the Fourier Transform reveals the following convolution relationship:
\begin{equation}
\label{eq:grappa}
	y_i = g_{i} \circledast \mathcal{F}\left(m\right)+\mathrm{noise}.
\end{equation}
Here $g_{i}$ is the Fourier Transform of the coil sensitivity pattern $S_{i}$ and $\circledast$ denotes the convolution operation. The  GRAPPA/SMASH formulation of parallel image reconstruction \citep{sodickson1997simultaneous,Griswold2002} involves filling in missing k-space data via combinations of acquired k-space data within a defined convolution kernel, prior to inverse Fourier transformation. 

Either formulation requires estimates of the coil sensitivity information in $S_{i}$ or $g_{i}$, which may be derived either from a separate reference scan or directly from the acquired undersampled k-space data itself. Reference scan methods are often used in the SENSE formulation, whereas GRAPPA formulations are typically self-calibrating, relying on subsets of fully-sampled data generally in central k-space regions.

The parallel imaging techniques described above may be combined productively with compressed sensing, via the use of sparsity-based regularizers.  
For example, one may extend Equation \ref{eq:CS} in Section \ref{sec:singlecoil_baseline} above to include multi-coil data as follows:
\begin{equation}
\label{eq:sense_cs}
\minimize_{m}\;\frac{1}{2} \sum_{i=1}^{n_c} \left\Vert \mathcal{P}\left(\mathcal{F}\left(S_{i}m\right)\right)-y_i\right\Vert _{2}^{2} + \lambda R\left(m\right).
\end{equation}
Various methods may be used to solve this reconstruction problem.  One frequently-used method is the ESPIRiT approach \citep{uecker2014espirit}, which harmonizes parallel imaging and compressed sensing in a unified framework.

As was the case for the classical single-coil baseline, the classical multi-coil baseline provided with the fastMRI dataset was adopted from the BART toolkit (Appendix \ref{appendix:bart-detail}). In the multi-coil case, the ESPIRiT algorithm was used to estimate coil sensitivities, and to perform parallel image reconstruction in combination with compressed sensing using a total-variation regularizer.  

\begin{table}[htb]
\centering
\begin{tabular}{cccccccc}\toprule
\multicolumn{8}{c}{Multi-coil classical baseline (TV model) applied to knee validation data} \\
\midrule
Acceleration & Regularization & \multicolumn{2}{c}{NMSE} & \multicolumn{2}{c}{PSNR} & \multicolumn{2}{c}{SSIM} \\ 
 \cmidrule(r){3-4} \cmidrule(r){5-6} \cmidrule(r){7-8} 
& & PD & PDFS & PD & PDFS & PD & PDFS \\ \midrule
4-fold & $10^{-4}$ & 0.0246 & 0.0972 & 31.6 & 27.4 & 0.677 & 0.53 \\
 & $10^{-3}$ & 0.0222 & \textbf{0.0951} & 32.1 & \textbf{27.5} & \textbf{0.693} & 0.554 \\
 & $10^{-2}$ & \textbf{0.0198} & 0.0971 & \textbf{32.6} & \textbf{27.5} & 0.675 & \textbf{0.588} \\
 & $10^{-1}$ & 0.0251 & 0.109 & 31.3 & 27 & 0.633 & 0.538 \\
 \midrule
8-fold & $10^{-4}$ & 0.0494 & 0.114 & 28.2 & 26.5 & 0.61 & 0.505 \\
 & $10^{-3}$ & 0.0447 & 0.112 & 28.6 & 26.6 & 0.626 & 0.524 \\
 & $10^{-2}$ & \textbf{0.0352} & \textbf{0.109} & \textbf{29.6} & \textbf{26.8} & \textbf{0.642} & \textbf{0.551} \\
 & $10^{-1}$ & 0.0389 & 0.114 & 29.2 & 26.7 & 0.632 & 0.527 \\
\bottomrule
\end{tabular} 
\caption{Validation set results for the classical baseline model with Total Variation regularization for the knee multi-coil task. Bold-faced numbers indicate the best performance for each image quality metric.}
\label{tbl:tv_multi}
\end{table}

\begin{table}[htb]
\centering
\begin{tabular}{ccrccc}\toprule
\multicolumn{6}{c}{Multi-coil classical baseline (TV model) applied to brain validation data} \\
\midrule
Acceleration & Regularization Weight & Sequence & NMSE & PSNR & SSIM \\ 
\midrule
 & $10^{-4}$ &
  AXT1 & 0.03971 & 31.63 & 0.5677 \\
& & AXT1POST & 0.02581 & 32.39 & 0.5814 \\
& & AXT2 & 0.03624 & 30.66 & 0.528 \\
& & AXFLAIR & 0.189 & 26.85 & 0.4512 \\
  \cmidrule{2-6}
& $10^{-3}$ &
  AXT1 & \textbf{0.03818} & \textbf{31.82} & \textbf{0.5724} \\
& & AXT1POST & 0.02353 & 32.81 & \textbf{0.5919}\\
& & AXT2 & 0.03457 & 30.86 & \textbf{0.5312} \\
4-fold & & AXFLAIR & 0.1869 & \textbf{26.96} & \textbf{0.4651} \\
\cmidrule{2-6}
& $10^{-2}$ &
  AXT1 & 0.03888 & 31.7 & 0.5376 \\
& & AXT1POST & \textbf{0.02199} & \textbf{33.17} & 0.5522 \\
& & AXT2 & \textbf{0.03419} & \textbf{30.9} & 0.4923 \\
& & AXFLAIR & \textbf{0.1886} & 26.75 & 0.4435 \\
\cmidrule{2-6}
& $10^{-1}$ &
  AXT1 & 0.04916 & 30.54 & 0.5193 \\
& & AXT1POST & 0.02956 & 31.84 & 0.5284 \\
& & AXT2 & 0.04708 & 29.39 & 0.4651 \\
& & AXFLAIR & 0.1934 & 26.14 & 0.4048 \\
\midrule
 & $10^{-4}$
  & AXT1 & 0.06911 & 29.01 & 0.4823 \\
& & AXT1POST & 0.05457 & 29.09 & 0.498 \\
& & AXT2 & 0.07904 & 27.05 & 0.4426 \\
& & AXFLAIR & 0.4421 & 23.93 & 0.3549 \\
\cmidrule{2-6}
& $10^{-3}$ &
   AXT1 & 0.06721 & 29.13 & 0.488 \\
& & AXT1POST & 0.05287 & 29.24 & 0.5039 \\
& & AXT2 & 0.078 & 27.11 & 0.4405 \\
8-fold & & AXFLAIR & \textbf{0.1869} & \textbf{26.96} & 0.4627 \\
\cmidrule{2-6}
& $10^{-2}$ &
  AXT1 & \textbf{0.05935} & \textbf{29.68} & 0.5145 \\
& & AXT1POST & \textbf{0.04514} & 29.92 & 0.5325 \\
& & AXT2 & \textbf{0.07486} & \textbf{27.29} & 0.4336 \\
& & AXFLAIR & 0.3893 & 24.15 & 0.3678 \\
\cmidrule{2-6}
& $10^{-1}$ &
   AXT1 & 0.06322 & 29.35 & \textbf{0.5928} \\
& & AXT1POST & 0.04904 & 29.54 & \textbf{0.6187} \\
& & AXT2 & 0.0874 & 26.6 & \textbf{0.495} \\
& & AXFLAIR & 0.2773 & 24.66 & \textbf{0.4726} \\
\bottomrule
\end{tabular} 
\caption{Validation set results for the classical baseline model with Total Variation regularization for the brain multi-coil task. Bold-faced numbers indicate the best performance for each image quality metric.}
\label{tbl:tv_brain_multi}
\end{table}

Results using this baseline model are summarized in Table \ref{tbl:tv_multi} and \ref{tbl:tv_brain_multi} . The experimental setup is identical to the single-coil scenario, except that we compare the reconstructions with the root-sum-of-squares ground truth instead of the ESC ground truth. 


\subsection{Single-coil Deep-Learning Baselines (knee only)}

\begin{figure}
\includegraphics[width=\textwidth,keepaspectratio]
{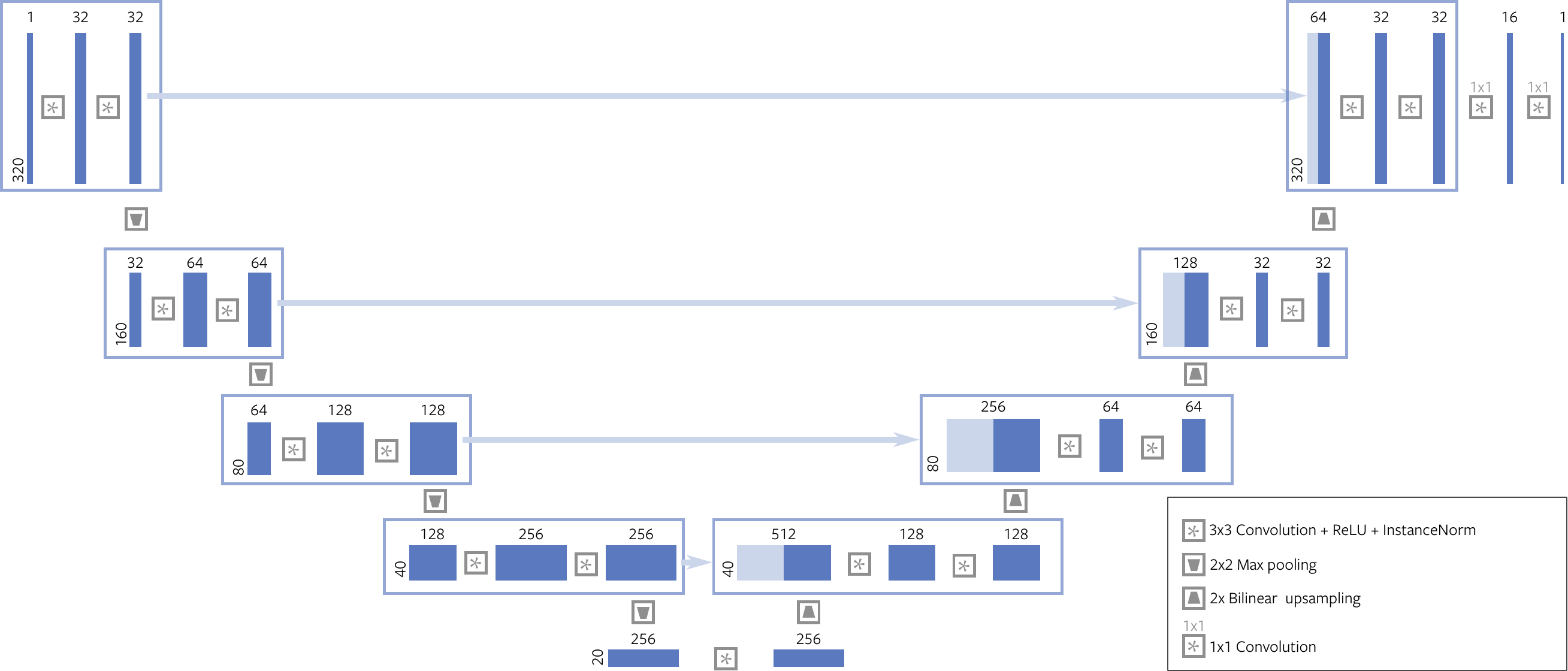}
\caption{\label{fig:baseline_unet} Single-coil baseline U-Net architecture}
\end{figure}

\label{sec:singlecoil_deep_baseline} Various deep-learning techniques based on Convolutional Neural Networks have recently been proposed to tackle the problem of reconstructing MR images from undersampled k-space data \citep{Hammernik2016,WangSYPZLFL16,varnet,SchlemperCHPR17,Zhu2018,Hyun2018DeepLF,Han2017}. Many of these proposed methods are based on the U-Net architecture introduced in \citep{DBLP:journals/corr/RonnebergerFB15}. U-Net models and their variants have successfully been used for many image-to-image prediction tasks including MRI reconstruction \citep{Hyun2018DeepLF,Han2017} and image segmentation \citep{DBLP:journals/corr/RonnebergerFB15}.

The U-Net single-coil baseline model included with the fastMRI data release (Figure~\ref{fig:baseline_unet}) consists of two deep convolutional networks, a down-sampling path followed by an up-sampling path. The down-sampling path consists of blocks of two 3$\times$3 convolutions each followed by instance normalization \citep{Ulyanov2016InstanceNT} and Rectified Linear Unit (ReLU) activation functions. The blocks are interleaved by down-sampling operations consisting of max-pooling layers with stride 2 which halve each spatial dimension. The up-sampling path consists of blocks with a similar structure to the down-sampling path, interleaved with bilinear up-sampling layers which double the resolution between blocks. Each block consists of two 3$\times$3 convolutions with instance normalization \citep{Ulyanov2016InstanceNT} and ReLU activation layers. In contrast to the down-sampling path, the up-sampling path concatenates two inputs to the first convolution in each block: the up-sampled activations from the previous block, together with the activations that follow the skip connection from the block in the down-sampling path with the same resolution (horizontal arrows in Figure~\ref{fig:baseline_unet}). At the end of the up-sampling path, we include a series of 1$\times$1 convolutions that reduce the number of channels to one without changing the spatial resolution.

For the single-coil MRI reconstruction case, the zero-filled image is used as the input to the model. The zero-filled image is obtained by first inserting zeros at the location of all unobserved k-space values, applying a two-dimensional Inverse Fourier Transform (IFT) to the result, and finally computing the absolute value. The result is center cropped to remove any readout and phase oversampling. Using the notation from section \ref{sec:singlecoil_baseline}, the zero-filled image is given by $\tilde{m} = \mathcal{C}(\left\vert \mathcal{F}^{-1}(\mathcal{P}(y)) \right\vert) $, where $\mathcal{C}$ is the linear operator corresponding to the center cropping and $\mathcal{F}^{-1}$ is the two-dimensional IFT.

The entire network is trained on the training data in an end-to-end manner to minimize the mean absolute error with respect to corresponding ground truth images. Let $B_\theta(m)$ be the function computed by the U-Net model, where $\theta$ represents the parameters of the model. Then the training process corresponds to the following optimization problem:
\begin{equation}
\minimize_{\theta}\; \frac{1}{2} \sum_{i=0}^{n_\text{data}} \left\Vert B_{\theta}(\tilde{m}^{(i)}) - m^{(i)} \right\Vert _{1},
\end{equation}
where the ground truths $m^{(i)}$ are obtained using the ESC method described in Section \ref{sec:esc}. Our particular single-coil U-Net baseline model was trained on 973 image volumes in the training set, using the RMSProp algorithm~\cite{rmsprop}. We used an initial learning rate of 0.001, which was multiplied by 0.1 after 40 epochs, after which the model was trained for an additional 10 epochs. During training, we randomly sampled a different mask for each training example in each epoch independently using the protocol described in Section \ref{sec:undersampling} for the test data. At the end of each epoch, we recorded the NMSE on the validation data. After training, we picked the model that achieved the lowest validation NMSE.

\begin{table}[htb]
\centering
\begin{tabular}{ccccccccc}\toprule
\multicolumn{9}{c}{Single-coil U-Net baseline applied to knee validation data} \\
\midrule
Acceleration & Channels & \#Params & \multicolumn{2}{c}{NMSE} & \multicolumn{2}{c}{PSNR} & \multicolumn{2}{c}{SSIM} \\ 
 \cmidrule(r){4-5} \cmidrule(r){6-7} \cmidrule(r){8-9} 
& & & PD & PDFS & PD & PDFS & PD & PDFS \\ \midrule
4-fold & 32 & 3.35M & 0.0161 & 0.0531 & 33.78 & 29.90 & 0.81 &     0.631 \\
   & 64 & 13.39M & 0.0157 & 0.0528 & 33.90 & 29.9 &0.813 &    0.633 \\
   & 128 & 53.54M & \textbf{0.0154} & \textbf{0.0525} & \textbf{34.01} & \textbf{29.95} & \textbf{0.815} &     0.634 \\
   & 256 & 214.16M & \textbf{0.0154} &    \textbf{0.0525} & 34.00 & \textbf{29.95} & \textbf{0.815} &     \textbf{0.636} \\
\midrule
8-fold & 32& 3.35M & 0.0283 & 0.0698 & 31.13 & 28.6 & 0.754 & 0.555\\
   & 64 & 13.39M & 0.0272 & 0.0693 & 31.30 &   28.63 & 0.758 & 0.558 \\
   & 128& 53.54M & 0.0265 & 0.0686 & 31.44 &   28.68 & 0.761 & 0.558 \\
   & 256& 214.16M & \textbf{0.0261} & \textbf{0.0682} & \textbf{31.5} & \textbf{28.71} & \textbf{0.762} & \textbf{0.559} \\
\bottomrule
\end{tabular} 
\caption{Validation set results for the U-Net baseline model trained for the single-coil task. The channels column  denotes the number of output channels of the first convolution in the model. Doubling this number of channels roughly quadruples the total number of parameters in the model.
Bold-faced numbers indicate the best performance for each image quality metric.}
\label{tbl:unet_single}
\end{table}

Table \ref{tbl:unet_single} presents the results from running trained U-Net models of different capacities on the single-coil validation data. These results indicate that the trained U-Net models perform significantly better than the classical baseline method.  The best U-Net models obtain 40-50\% relative improvement over the classical methods (see Table \ref{tbl:tv_single}) in terms of NMSE.

The performance of the U-Net models continues to increase with increasing model capacity, and even the largest model with over 200 million parameters is unable to overfit the training data. These improvements begin to saturate after 50 million parameters for the simpler 4-fold acceleration case. However, for the more challenging 8-fold acceleration task, the largest model performs significantly better than the smaller models. This suggests that models with very large capacities trained on large amounts of data can enable high acceleration factors.

\begin{table}[htb]
\centering
\begin{tabular}{ccccc}\toprule
\multicolumn{5}{c}{Single-coil classical and U-Net baselines applied to test data} \\
\midrule
Model & Acceleration & NMSE & PSNR & SSIM \\ 
\midrule
Classical Model (Total Variation)
	& 4-fold & 0.0479 & 30.69 & 0.603 \\
    & 8-fold & 0.0795 & 27.12 & 0.469 \\
    & Aggregate& 0.0648 & 28.77 & 0.531 \\
\midrule
U-Net & 4-fold & 0.0320 & 32.22 & 0.754 \\
	  & 8-fold & 0.0480 & 29.45 & 0.651 \\
      & Aggregate& 0.0406 & 30.7 & 0.699 \\
\bottomrule
\end{tabular}
\caption{\label{tbl:test_res_single} Comparison of classical and U-Net baseline performance for the single-coil task with test data}
\end{table}

Table \ref{tbl:test_res_single} compares the performance of the classical and the U-Net baseline models for the single-coil task, as applied to the test dataset. For the classical baseline model, we chose the best regularization weights for each modality and for each acceleration factor based on the validation data results, resulting in a regularization weight of 0.1 for 8-fold acceleration on Proton Density without fat suppression and 0.01 for every other case.  For the U-Net baseline model, we chose the model with the largest capacity.

\subsection{Multi-coil Deep-Learning Baselines}
\label{sec:multicoil_deep_baseline} In the multi-coil MRI reconstruction task, we have one set of undersampled k-space measurements from each coil, and a different zero-filled image can be computed from each coil. These coil images can be combined using the root-sum-of-squares algorithm. Let $\tilde{m}_i$ be the zero-filled image from coil $i$. With \begin{math} \tilde{m}_{\text{rss}} \end{math} defined as in Equation \ref{eq:RSS}, the U-Net model described in Section \ref{sec:singlecoil_deep_baseline} can be used for the multi-coil reconstruction task by simply feeding this combined image in as input: $B_\theta(\tilde{m}_{rss}).$ The model is trained to minimize the mean absolute error loss similarly to the single-coil task. The training procedure is also identical to the single-coil case except that the root-sum-of-squares image is used as the ground truth as described in Section \ref{sec:ground-truth}.


\begin{table}[htb]
\centering
\begin{tabular}{ccccccccc}\toprule
\multicolumn{9}{c}{Multi-coil U-Net baseline applied to knee validation data} \\
\midrule
Acceleration & Channels & \#Params & \multicolumn{2}{c}{NMSE} & \multicolumn{2}{c}{PSNR} & \multicolumn{2}{c}{SSIM} \\ 
 \cmidrule(r){4-5} \cmidrule(r){6-7} \cmidrule(r){8-9} 
& & & PD & PDFS & PD & PDFS & PD & PDFS \\ \midrule
4-fold & 32 & 3.35M & 0.0066 &    0.0122 & 36.7 &    35.97 & 0.9192 &     0.8595\\
   & 64 & 13.39M & 0.0063 &    0.0120 & 36.95 &      36.11 & 0.9224 &     0.8615\\
   & 128 & 53.54M & 0.0057 &    0.0113 & 37.38 &      36.33 & 0.9266 &     0.8641 \\
   & 256 & 214.16M & \textbf{0.0054} & \textbf{0.0112} & \textbf{37.58} &  \textbf{36.39} & \textbf{0.9287} &\textbf{0.8655} \\
\midrule
8-fold & 32 & 3.35M & 0.0144 &    0.0197 & 33.31 &      33.82 & 0.8778 &     0.8213\\
   & 64 & 13.39M & 0.0136 &   0.0198 & 33.56 &     33.93 & 0.8825 &    0.8238\\
   & 128 & 53.54M & 0.0123 &    0.0179 & 34.01 &      \textbf{34.25} & 0.8892 &     0.8277 \\
   & 256 & 214.16M & \textbf{0.0120} & \textbf{0.0181} & \textbf{34.12} &      34.23 & \textbf{0.8915} & \textbf{0.8286} \\
\bottomrule
\end{tabular} 
\caption{Validation set results for the U-Net baseline model trained for the multi-coil task. Bold-faced numbers indicate the best performance for each image quality metric.}
\label{tbl:unet_multi}
\end{table}

\begin{table}[htb]
\centering
\begin{tabular}{cccrccc}\toprule
\multicolumn{7}{c}{Multi-coil U-Net baseline applied to brain validation data} \\
\midrule
Acceleration & Channels & \#Params & Sequence & NMSE & PSNR & SSIM \\ 
\midrule
        & 32 & 3.35M & AXT1 & 0.01498 & 35.67 & 0.9215 \\
        & & & AXT1POST & 0.013 & 35.43 & 0.9298 \\
        & & & AXT2 & 0.02249 & 32.51 & 0.9112 \\
        & & & AXFLAIR & 0.1572 & 30.73 & 0.7869 \\
\cmidrule{2-7}
        & 64 & 13.39M & AXT1 & 0.01571 & 35.57 & 0.922 \\
        & & & AXT1POST & 0.01313 & 35.41 & 0.9307 \\
        & & & AXT2 & 0.02014 & 32.98 & 0.9151 \\
4-fold  & & & AXFLAIR & 0.1579& 30.96 & 0.7917 \\
\cmidrule{2-7}
        & 128 & 53.54M & AXT1 & 0.0142 & 35.92 & 0.9243 \\
        & & & AXT1POST & 0.01231 & 35.69 & 0.9332 \\
        & & & AXT2 & 0.01855 & 33.34 & 0.9175 \\
        & & & AXFLAIR & 0.1566 & 30.98 & 0.7932 \\
\cmidrule{2-7}
        & 256 & 214.16M & AXT1 & \textbf{0.01317} & \textbf{36.24} & \textbf{0.9275} \\
        & & & AXT1POST & \textbf{0.0111} & \textbf{36.11} & \textbf{0.9361} \\
        & & & AXT2 & \textbf{0.01733} & \textbf{33.63} & \textbf{0.9207} \\
        & & & AXFLAIR & \textbf{0.1532} & \textbf{31.52} & \textbf{0.7985} \\
\midrule
        & 32 & 3.35M & AXT1 & 0.04289 & 31.5 & \textbf{0.8885} \\
        & & & AXT1POST & 0.04186 & 31.71 & 0.8816 \\
        & & & AXT2 & 0.04357 & 30.86 & 0.8759 \\
        & & & AXFLAIR & \textbf{0.1594} & \textbf{32.86} & \textbf{0.8188}\\
\cmidrule{2-7}
        & 64 & 13.39M & AXT1 & \textbf{0.04205} & \textbf{32.56} & 0.8876 \\
        & & & AXT1POST & 0.04034 & 31.89 & 0.883 \\
        & & & AXT2 & 0.04248 & 31.1 & 0.8753\\
8-fold  & & & AXFLAIR & 0.1818 & 30.49 & 0.7843 \\
\cmidrule{2-7}
        & 128 & 53.54M & AXT1 & 0.04706 & 31.82 & 0.8804 \\
        & & & AXT1POST & 0.04005 & 31.47 & 0.8828 \\
        & & & AXT2 & 0.04311 & 30.13 & 0.8806 \\
        & & & AXFLAIR & 0.2 & 28.97 & 0.7779 \\
\cmidrule{2-7}
        & 256 & 214.16M & AXT1 & 0.0443 & 32.02 & 0.8837 \\
        & & & AXT1POST & \textbf{0.04028} & \textbf{31.95} & \textbf{0.8845} \\
        & & & AXT2 & \textbf{0.04167} & \textbf{31.29} & \textbf{0.8811} \\
        & & & AXFLAIR & 0.1565 & 30.49 & 0.7805 \\
\bottomrule
\end{tabular} 
\caption{Validation set results for the U-Net baseline model trained for the brain multi-coil task. Bold-faced numbers indicate the best performance for each image quality metric.}
\label{tbl:unet_brain_multi}
\end{table}

As is the case for the single-coil task, the multi-coil U-Net baselines substantially outperform the classical baseline models (compare Table \ref{tbl:unet_multi} and \ref{tbl:unet_brain_multi} with Table \ref{tbl:tv_multi} and \ref{tbl:tv_brain_multi}). Note that this is true despite the fact that the multi-coil U-Net baseline defined above does \emph{not} take coil sensitivity information into account, and therefore neither includes a direct parallel image reconstruction nor accounts for sparsity or other correlations among coils. Models that incorporate coil sensitivity information are expected to perform better than the current multi-coil U-Net baselines.  Table~\ref{tbl:unet_multi} and Table~\ref{tbl:unet_brain_multi} shows, once again, that the performance of the U-Net models improves with model size, with the largest U-Net baseline model providing the best performance.

\begin{table}[htb]
\centering
\begin{tabular}{cccccc}\toprule
\multicolumn{5}{c}{Multi-coil classical and U-Net baselines applied to test data} \\
\midrule
Dataset & Model & Acceleration & NMSE & PSNR & SSIM \\ 
\midrule
Knee & Classical Model (Total Variation) 
    & 4-fold & 0.0503 & 30.88 & 0.628 \\
    & & 8-fold & 0.0760 & 28.25 & 0.593 \\
    & & Aggregate& 0.0633 & 29.54 & 0.610 \\
& U-Net & 4-fold & 0.0106 & 35.91 & 0.904 \\
	  & & 8-fold & 0.0171 & 33.57 & 0.858 \\
      & & Aggregate& 0.0139 & 34.7 & 0.881\\
\midrule
Brain & Classical Model (Total Variation)
      & 4-fold & 0.1388 & 27.53 & 0.4439\\
	  & & 8-fold & 0.03753 & 31.32 & 0.5135\\
      & & Aggregate & 0.0882 & 29.42 & 0.4787\\
& U-Net & 4-fold & 0.0107 & 38.13 & 0.9446\\
	  & & 8-fold & 0.0233 & 34.52 & 0.9146\\
      & & Aggregate & 0.017 & 36.325 & 0.9296\\

\bottomrule
\end{tabular}
\caption{\label{tbl:test_res_multi} Comparison of classical and U-Net baseline performance for the multi-coil task with knee test data.}
\end{table}

Table \ref{tbl:test_res_multi} compares the performance of the classical and the U-Net baseline models for the multi-coil task, as applied to the test dataset. For the classical baseline model, we chose the best regularization weights for each modality and for each acceleration factor based on the validation data results. For knees this resulted in a regularization weight of 0.001 for 4-fold undersampling for Proton Density with Fat Suppression and 0.01 for every other acquisition type. For brain this resulted in a regularization weight of 0.001 for 8-fold AXFLAIR and 4-fold AXT1, and 0.01 for every other acquisition type. For the U-Net baseline model, we chose the model with the largest capacity.

\begin{figure}
\centering
\includegraphics[width=0.49\textwidth,keepaspectratio]{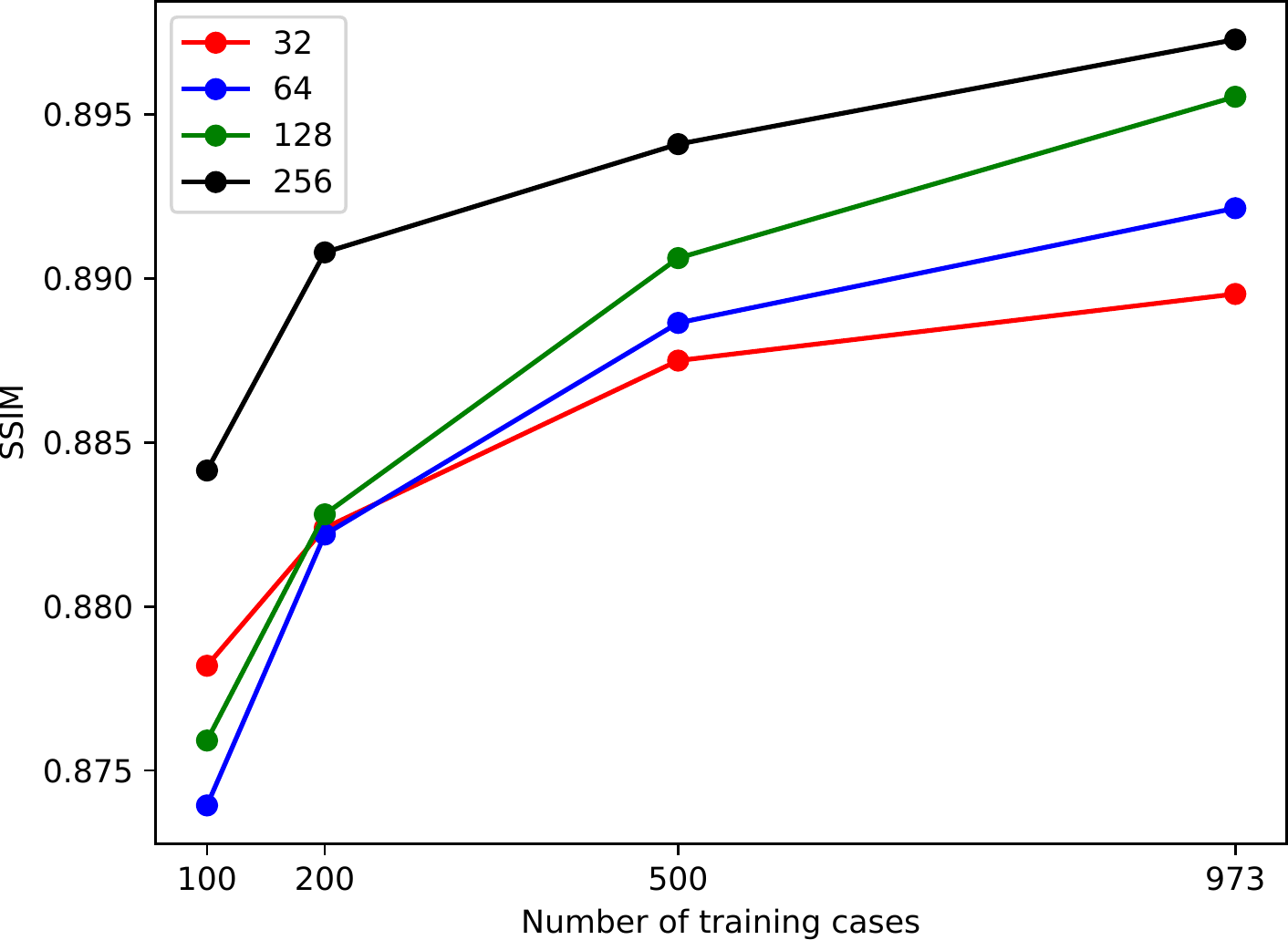}
\includegraphics[width=0.49\textwidth,keepaspectratio]{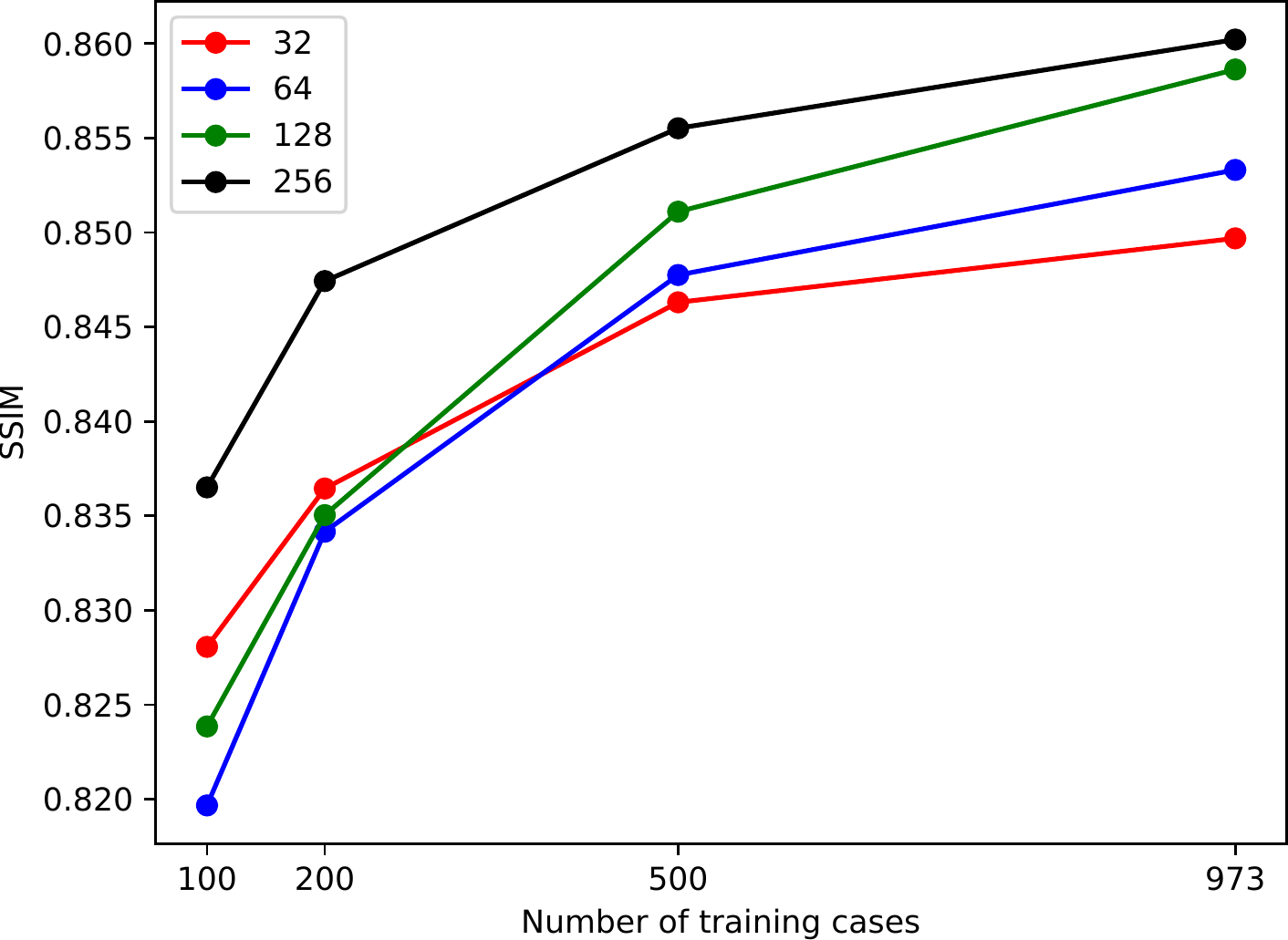}
\caption{\label{fig:ssim_data_size} Results from training the U-Net on different amounts of training data for the multi-coil knee challenge with 4-fold acceleration (left) and 8-fold acceleration (right). Each line represents a model with a different number of channels.}
\end{figure}

To appreciate the value of the dataset size, we study how model performance scales with the amount of data used to train a model. To this end, we trained several U-Net models with varying model capacities on different sized subsets of the training data. Figure~\ref{fig:ssim_data_size} shows the SSIM metric computed on the validation data for the multi-coil task. It is evident from these results that training with larger amounts of data yields substantial improvements in the quality of reconstructions, which highlights the need for the release of large datasets like fastMRI.

As mentioned in Section \ref{sec:dicom_data}, the fastMRI dataset also includes a large set of DICOM images that can be used as additional training data. It is possible that the baseline U-Net models could be improved further by making use of this additional data.

\begin{figure}
\centering
\includegraphics[width=\textwidth,keepaspectratio]{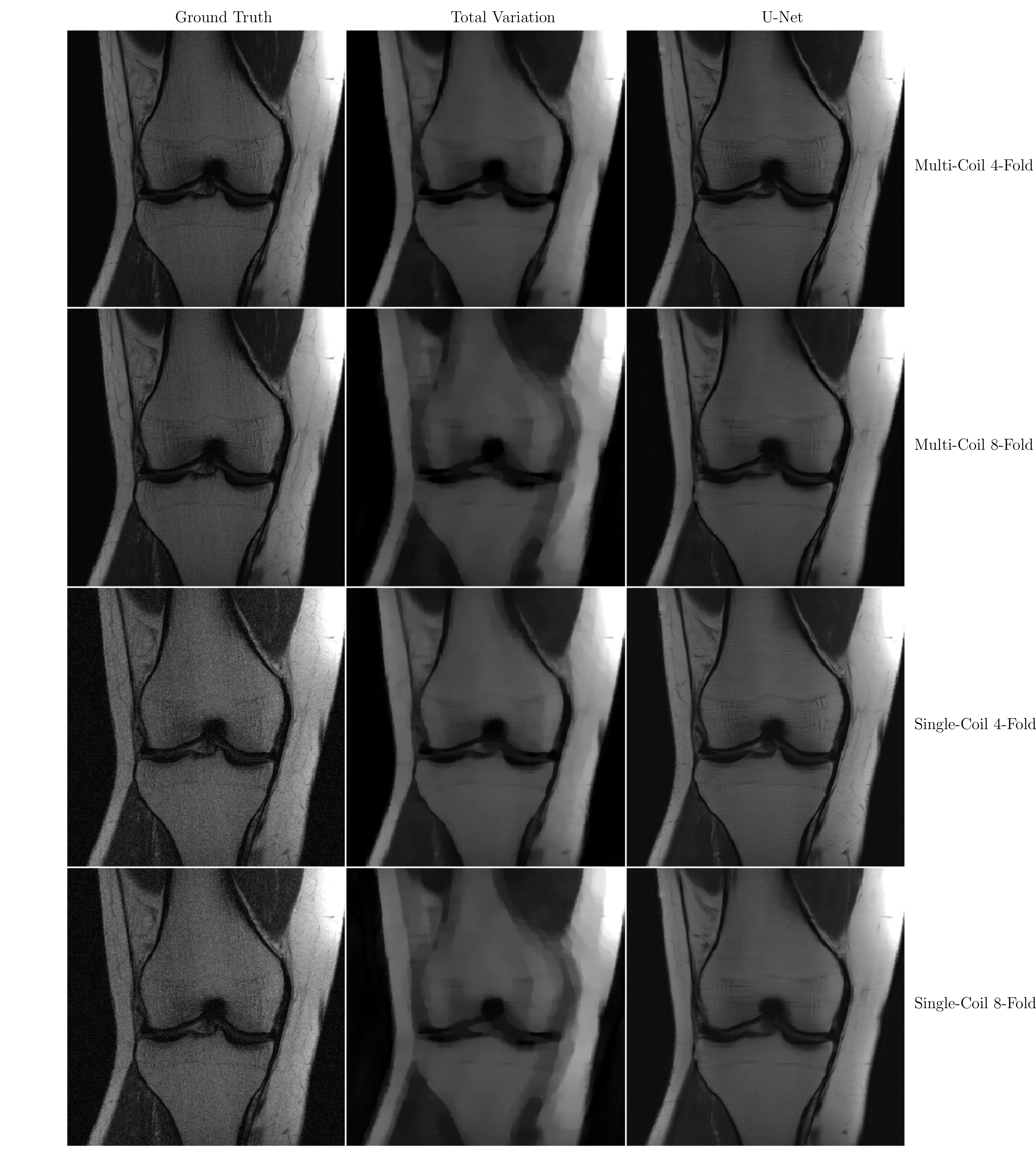}
\caption{\label{fig:example-pd} Example knee reconstructions}
\end{figure}

\begin{figure}
\centering
\includegraphics[width=\textwidth,keepaspectratio]{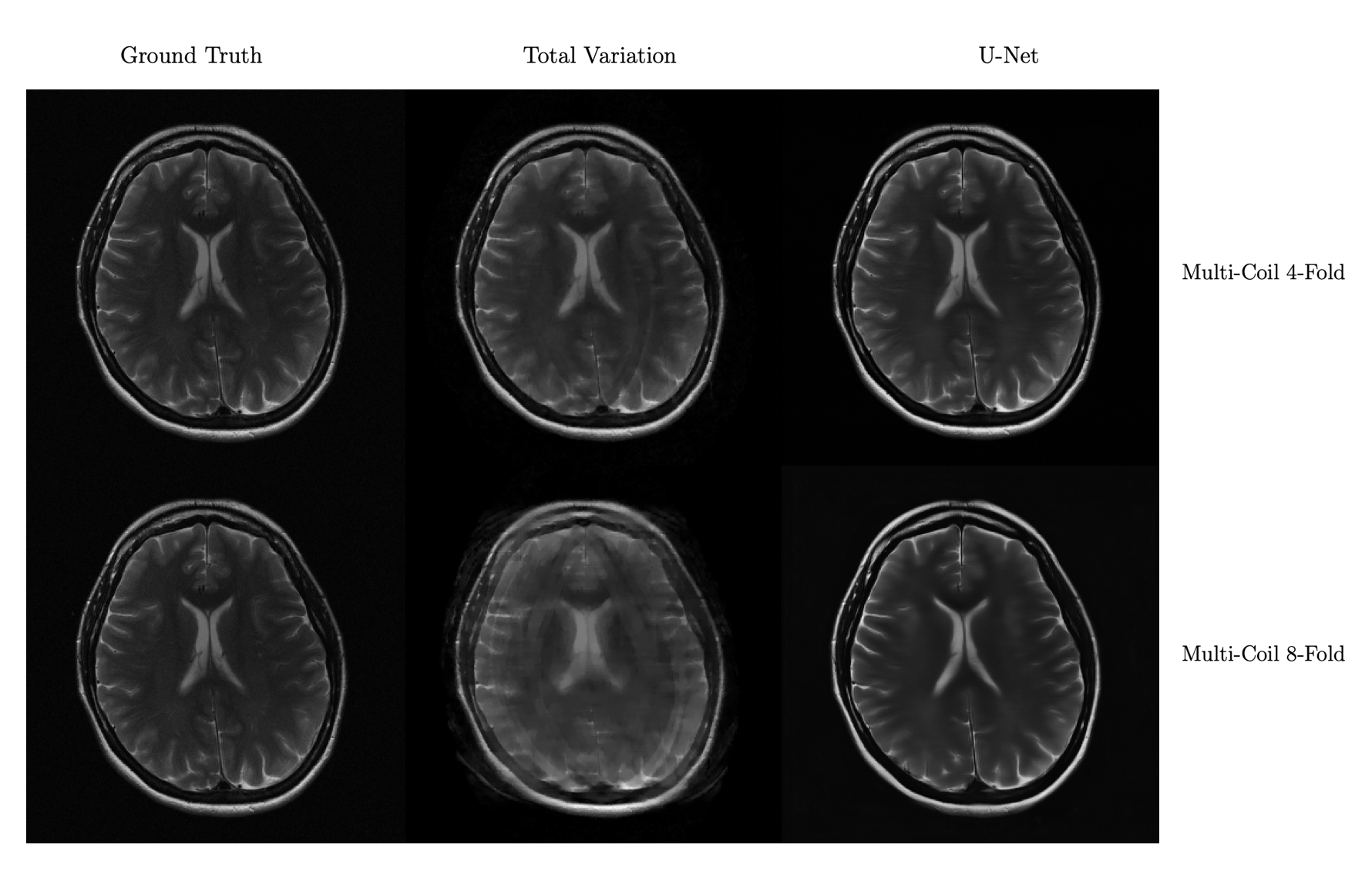}
\caption{\label{fig:example-brain} Example brain reconstructions}
\end{figure}

\section{Discussion}
\label{sec:discussion} MR image reconstruction is an inverse problem, and thus it has many connections to inverse problems in the computer vision literature \cite{szeliski2011computer, Fan-1712-00202, ChangLPKS17, Ulyanov-1711-10925}, such as super-resolution, denoising and in-painting. In all of these inverse problems, the goal is to recover a high-dimensional ground truth image from a lower-dimensional measurement. Such ill-posed problems are very difficult to solve since there exists an infinite number of high-dimensional images that can result in the same-low dimensional measurement. In order to simplify the problem, an assumption is often made that only a small number of high-resolution images would correspond to natural images \cite{ChangLPKS17}. Given that MRI reconstruction is a similar inverse problem, we hope that the computer vision community, as well as the medical imaging community, will find our dataset beneficial.

In the clinical setting, radiologists use MRI to search for abnormalities, make diagnoses, and recommend treatment options. Thus, contrary to many computer vision problems where small texture changes might not necessarily alter the overall satisfaction of the observer, in MRI reconstruction, extra care should be taken to ensure that the human interpreter is not misled by a very plausible but not necessarily correct reconstruction. This is especially important as image generation techniques increase in their ability to generate photo-realistic results \citep{pix2pixhd}. Therefore some research effort should be devoted to look for solutions that, by design, ensure correct diagnosis, and we hope that our dataset will provide a testbed for new ideas in these directions as well.

An important question in MRI reconstruction is the choice of the evaluation metric. The current consensus in the MRI community is that global metrics, such as NMSE, SSIM and PSNR, do not necessarily capture the level of detail required for proper evaluation of MRI reconstruction algorithms \cite{MIAO2013596, Huo2006}. A natural question arises: what would the optimal metric be? An ideal MRI reconstruction algorithm should produce sharp, trustworthy images, that ultimately ensure the proper radiologic interpretation. While our dataset will help ensure consistent evaluation, we hope that it will also trigger research on MRI reconstruction metrics. This goal will be impossible to achieve without clinical studies involving radiologists evaluating fully-sampled and undersampled MRI reconstructions to make sure that both images lead to the same diagnosis.

Although this dataset provides an excellent entry point for machine learning methods for MR reconstruction, there are some aspects of MR imaging that we have not yet considered here. Physical effects such as spin relaxation, eddy currents and field distortions are not at present explicitly accounted for in our retrospective undersampling approaches or our baseline models. The manifestation of these effects depends upon the object being imaged, the MRI scanner used, and even the sampling pattern selected. Extending the results from methods developed for this challenge to the clinic remains an open problem, but we believe the provision of this dataset is an important first step on the path to this goal.

\section{Conclusion}
\label{sec:conclusion}
In this work we detailed the fastMRI dataset: the largest raw MRI dataset to be made publicly available to date. Previous public datasets have focused on post-processed magnitude images for specific biologic and pathologic questions. Although our dataset was originally acquired for a focused task, the inclusion of raw k-space data allows methods to be developed for the imaging pipeline itself, in principle allowing them to be applied on any MRI scanner for any imaging task.

In addition to the data, we provide evaluation metrics and baseline algorithms to aid the research community in assessing new approaches. Consistent evaluation of MRI reconstruction techniques is provided by a leaderboard using held-out test data. 

We hope that the availability of this dataset will accelerate research in MR image reconstruction, and will serve as a benchmark during training and validation of new algorithms.

\section{Acknowledgements}
\label{sec:acknowledgements}
We acknowledge grant support from the National Institutes of Health under grants NIH R01 EB024532 and NIH P41 EB017183. We would also like to thank Michela Paganini and Mark Tygert.

\section{Changelog}
\begin{itemize}
    \item V1 (Nov 2018): Arxiv preprint describing the fastMRI knee dataset.
    \item V2 (Dec 2019): Added neuro dataset.
\end{itemize}

\clearpage

{\small
\bibliographystyle{plainnat}
\bibliography{mri}
}

\appendix
\section{Raw k-space File Descriptions}
ISMRMRD files were converted into simpler HDF5 files that store the entire k-space in a single tensor. One HDF5 file was created per volume. The HDF5 files share the following common attributes:
\begin{description}
\item[acquisition] Acquisition protocol. For knee images this is either CORPD or CORPDF, indicating coronal proton density with or without fat saturation, respectively (see Figure~\ref{fig:acquisition}). For Brain images this is AXFLAIR, AXT1, AXT1POST or AXT2 (see Figure~\ref{fig:brain_acquisition}).
\item[ismrmrd\_header] The XML header copied verbatim from the ISMRMRD file that was used to generate the HDF5 file. It contains information about the scanner, field of view, dimensions of k-space, and sequence parameters.
\item[patient\_id] A unique string identifying the examination, and substituting anonymously for the patient identification.
\item[norm, max] The Euclidean norm and the largest entry of the target volume. For the multi-coil track the target volume is stored in reconstruction\_rss. For the single-coil track the target volume is stored in reconstruction\_esc. These two attributes are only available in the training and validation datasets.
\item[acceleration] Acceleration factor of the undersampled k-space trajectory (either 4 or 8). This attribute is only available in the test dataset.
\item[num\_low\_frequency] The number of low-frequency k-space lines in the undersampled k-space trajectory. This attribute is only available in the test dataset.
\end{description}
The rest of this section describes the format of the HDF5 files for the multi-coil and single-coil tracks.

\subsection{Multi-coil Track}
\begin{description}
\item[\{knee,brain\}\_multicoil\_train.tar.gz] Training dataset for the multi-coil track. The HDF5 files contain the following tensors:

\begin{description}
\item[kspace] Multi-coil k-space data. The shape of the kspace tensor is (number of slices, number of coils, height, width).

\item[reconstruction\_rss] root-sum-of-squares reconstruction of the multi-coil k-space data. The shape of the reconstruction\_rss  tensor is (number of slices, r\_height, r\_width). For knee images, height and width have been cropped to 320 x 320.

\end{description}

\item[\{knee,brain\}\_multicoil\_val.tar.gz] Validation dataset for the multi-coil track. The
HDF5 files have the same structure as the HDF5 files in
multicoil\_train.tar.gz.

\item[\{knee,brain\}\_multicoil\_test.tar.gz] Test dataset for the multi-coil track. The HDF5
files contain the following tensors:

\begin{description}
\item[kspace] Undersampled multi-coil k-space. The shape of the kspace tensor is
(number of slices, number of coils, height, width).

\item[mask] Defines the undersampled Cartesian k-space trajectory. The number of elements in the mask tensor is the same as the width of k-space.

\end{description}
\end{description}

\subsection{Single-coil Track (knee only)}

\begin{description}
\item[knee\_singlecoil\_train.tar.gz] Training dataset for the single-coil track. Note that only the knee dataset has a single-coil track. The HDF5 files contain the following tensors:

\begin{description}
\item[kspace] Emulated single-coil k-space data. The shape of the kspace tensor is (number of slices, height, width).

\item[reconstruction\_rss] root-sum-of-squares reconstruction of the multi-coil k-space that was used to derive the emulated single-coil k-space cropped to the center $320 \times 320$ region. The shape of the reconstruction\_rss tensor is (number of slices, 320, 320).

\item[reconstruction\_esc] The inverse Fourier transform of the single-coil k-space data cropped to the center $320 \times 320$ region. The shape of the reconstruction\_esc tensor is (number of slices, 320, 320).

\end{description}

\item[knee\_singlecoil\_val.tar.gz] Validation dataset for the single-coil track. The HDF5 files have the same structure as the HDF5 files in singlecoil\_train.tar.gz.

\item[knee\_singlecoil\_test.tar.gz] Test dataset for the single-coil track. Note that only the knee dataset has a single-coil track. The HDF5 files contain the following tensors:

\begin{description}
\item[kspace] Undersampled emulated single-coil k-space. The shape of the kspace tensor is (number of slices, height, width).

\item[mask] Defines the undersampled Cartesian k-space trajectory. The number of elements in the mask tensor is the same as the width of k-space.

\end{description}
\end{description}

\section{Classical Reconstruction with BART}
\label{appendix:bart-detail} The Berkeley Advanced Reconstruction Toolbox (BART) \cite{BART} \footnote{Version 0.4.03 \url{https://mrirecon.github.io/bart/}} contains implementations of standard methods for coil sensitivity estimation and undersampled MR image reconstruction incorporating parallel imaging and compressed sensing. We used this tool to produce the classical baseline MSE estimates, as well as the illustrations in Figure~\ref{fig:multicoil-example}. In this section we provide a brief introduction to the tool sufficient for reproducing our baseline results. We will use as an example a 640x368 undersampled MRI scan with 15 coils. The target region is a $320\times 320$ central region which will be cropped to after reconstruction.

BART provides a command line interface which acts on files in a simple storage format. Each multidimensional array is stored in a pair of files, a header file \verb|.hdr| and a data file \verb|.cfl|. The header file contains the dimensions of the array given in ASCII. In our running example, this should be \verb|input.hdr|:
\begin{quote}
    \verb|1 640 368 15|
\end{quote}
The CFL file contains the raw data in column-major order, stored as complex float values. Missing k-space values are indicated by 0 entries. BART provides Python and MATLAB interfaces for reading and writing this format.

When working with k-space data with BART, it is simplest to use data in "centered" form, where the low frequency values are in the center of the image, and the high frequency values are at the edges. Most FFT libraries output the data in uncentered form. BART provides a tool for conversion:
\begin{quote}
	\verb|bart fftshift 7 input output|
\end{quote}
The input and output are specified without file extensions. The value 7 above is a bitmask indicating the image is stored in axis 0,1,2 (1+2+4) of the input array. This bitmask is used in the commands that follow also. 

Uncentered k-space data is easily identified by comparing the magnitude of the corners versus the center of the array. Centered FFTs of natural data will have the largest magnitudes near the center of the array when plotted.

Parallel MR imaging is often performed as a two-step process consisting of coil-sensitivity estimation, then reconstruction assuming the estimated sensitivity maps are exact. BART implements this approach through the \verb|ecalib| and \verb|pics| commands. The coil-sensitivity maps can be estimated using the ESPIRiT approach using the command
\begin{quote}
\verb|bart ecalib | \\
\verb|    -m1      | \doc{Produce a single set of sensitivity maps} \\
\verb|    -r26     | \doc{Number of fully sampled reference lines} \\
\verb|    input output_sens |
\end{quote}
The central reference region is used by BART to estimate the coil sensitivities. This area is also known as the auto-calibration region. The number of lines used in our masking procedure is a percentage of the k-space width, as described in Section \ref{sec:raw-multi-coil-description}.

Given the estimated coil sensitivities, a reconstruction using TV regularization can be performed with
\begin{quote}
\verb|bart pics | \\
\verb|    -d4            | \doc{Debug log level, use 0 for no stdout output} \\
\verb|    -i200          | \doc{Optimization iterations} \\
\verb|    -R T:7:0:0.05  | \doc{Use TV (T) with regularizer strength 0.05, with bitmask 7} \\
\verb|    input output_sens output |
\end{quote}
The output of this command is in CFL format. It can be converted to a PNG using \verb|bart toimg|. When using L1 wavelet regularization, the character "W" should be used in the \verb|R| option, with the additional \verb|-m| argument to ensure that ADMM is used.

       
\end{document}